\documentclass{vgtc}                          % final (conference style)
\ifpdf%                                % if we use pdflatex
  \pdfoutput=1\relax                   % create PDFs from pdfLaTeX
  \usepackage{graphicx}                % allow us to embed graphics files
  \DeclareGraphicsExtensions{.pdf,.png,.jpg,.jpeg} % for pdflatex we expect .pdf, .png, or .jpg files
\else%                                 % else we use pure latex
  \usepackage{graphicx}                % allow us to embed graphics files
  \DeclareGraphicsExtensions{.eps}     % for pure latex we expect eps files
\fi%

\graphicspath{{figures/}{pictures/}{images/}{./}} % where to search for the images

\usepackage{microtype}                 % use micro-typography (slightly more compact, better to read)
\PassOptionsToPackage{warn}{textcomp}  % to address font issues with \textrightarrow
\usepackage{textcomp}                  % use better special symbols
\usepackage{mathptmx}                  % use matching math font
\usepackage{times}                     % we use Times as the main font
\usepackage{cite}                      % needed to automatically sort the references
\usepackage{graphicx}
\usepackage{epsfig} % for postscript graphics files
\usepackage{times} % assumes new font selection scheme installed
\usepackage{amsmath} % assumes amsmath package installed
\usepackage{amssymb}  % assumes amsmath package installed
\usepackage{hyperref}
\usepackage{subfig}
\usepackage{pbox}
\usepackage{multirow}
\usepackage{hyperref}
\graphicspath{ {figures/} }
\newcommand{\methodtitle}{\mbox{MaskFusion}}

\newcommand{\specialcell}[2][c]{%
  \begin{tabular}[#1]{@{}c@{}}#2\end{tabular}}

\newcommand{\tablefootnote}[1]{}
\onlineid{0}

\vgtccategory{Research}

\vgtcinsertpkg

 \title{\methodtitle{}: Real-Time Recognition, Tracking and Reconstruction \\of Multiple Moving Objects}
\author{Martin R{\"u}nz\thanks{e-mail: martin.runz.15@ucl.ac.uk} %
\and Maud Buffier\thanks{e-mail: maud.buffier@gmail.com} %
\and Lourdes Agapito\thanks{e-mail:l.agapito@ucl.ac.uk} \phantom{\thanks{\url{http://visual.cs.ucl.ac.uk/pubs/maskfusion/}}}}
\affiliation{\scriptsize Department of Computer Science \\ University College London, UK}

\teaser{
  \centering
  \subfloat[Frame 400]{\includegraphics[width=0.33 \linewidth]{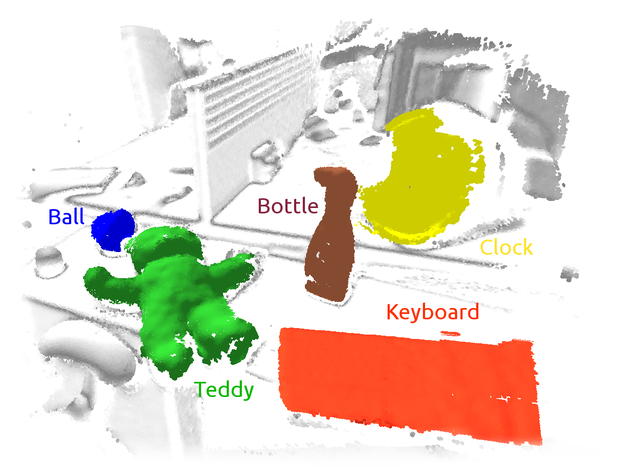}}
  \subfloat[Frame 700]{\includegraphics[width=0.33 \linewidth]{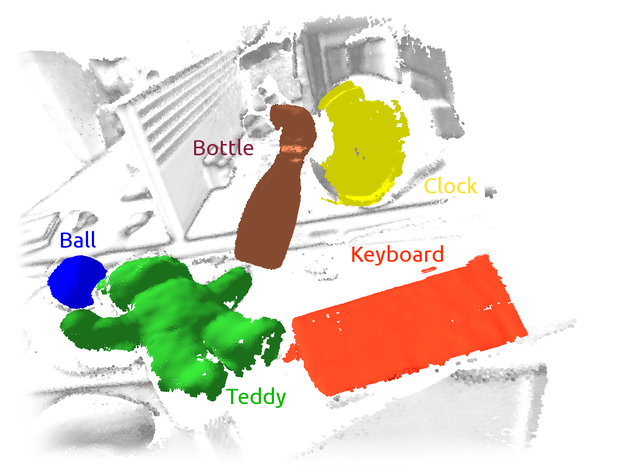}}
  \subfloat[Frame 900]{\includegraphics[width=0.33 \linewidth]{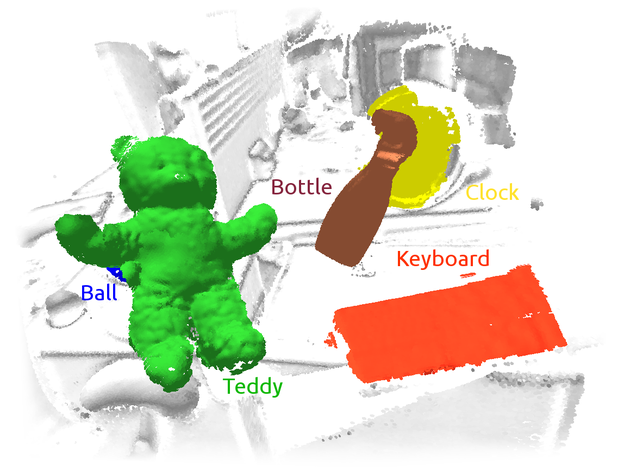}}
  \\
  \subfloat{\includegraphics[width=0.16 \linewidth]{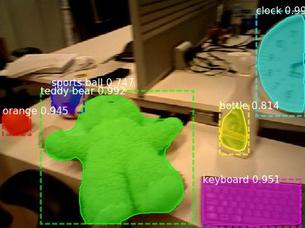}}
  \subfloat{\includegraphics[width=0.16 \linewidth]{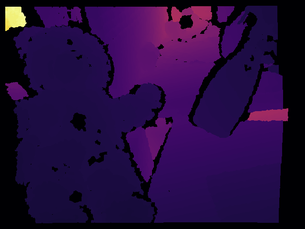}}~
  \subfloat{\includegraphics[width=0.16 \linewidth]{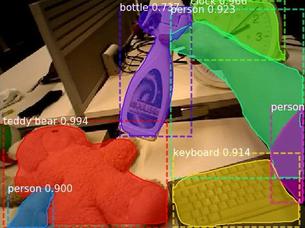}}
  \subfloat{\includegraphics[width=0.16 \linewidth]{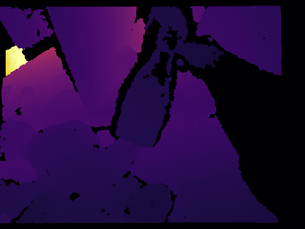}}~
  \subfloat{\includegraphics[width=0.16 \linewidth]{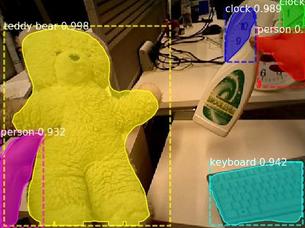}}
  \subfloat{\includegraphics[width=0.16 \linewidth]{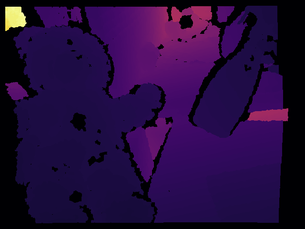}}

  \caption{A series of 3 frames illustrating the recognition, tracking and mapping capabilities of \methodtitle{}. The first row highlights the system's output: A reconstruction of the background (white), keyboard (orange), clock (yellow), sports ball (blue),  teddy-bear (green) and spray-bottle (brown). While the camera was in motion during the whole sequence, the bottle and the teddy started moving from frame 500 and 690 onwards, respectively. Note that \methodtitle{} explicitly avoided to reconstruct geometry related to the person holding the objects. The second row shows the input RGBD frames and semantic masks produced by the segmentation neural network as an overlay.
 }
}

\abstract{We present \methodtitle{}, a \emph{real-time, object-aware,
    semantic and dynamic RGB-D SLAM system} that goes beyond
  traditional systems which output a purely geometric map of a static
  scene. \methodtitle{} recognizes, segments and assigns semantic
  class labels to different objects in the scene, while tracking and
  reconstructing them even when they move independently from the
  camera. As an RGB-D camera scans a cluttered scene, image-based
  instance-level semantic segmentation creates semantic object masks
  that enable real-time object recognition and the creation of an
  object-level representation for the world map. Unlike previous
  recognition-based SLAM systems, \methodtitle{} does not require
  known models of the objects it can recognize, and can deal with
  multiple independent motions.  \methodtitle{} takes full advantage
  of using instance-level semantic segmentation to enable semantic
  labels to be fused into an object-aware map, unlike recent semantics
  enabled SLAM systems that perform voxel-level semantic
  segmentation. We show augmented-reality applications that
  demonstrate the unique features of the map output by \methodtitle{}:
  instance-aware, semantic~and~dynamic. Code will be made
  available\textsuperscript{\textdaggerdbl}.}

\CCScatlist{
  \CCScatTwelve{Visual SLAM}{SLAM}{Visu\-al\-iza\-tion}{Tracking};
  \CCScatTwelve{Mapping}{Fusion}{RGBD}{Multi-object}
  \CCScatTwelve{Recognition}{Context}{Semantic}{Detection}
  \CCScatTwelve{Real-time}{Augmented-Reality}{Robotics}{}
}

\begin{document}


\firstsection{Introduction}
\maketitle

Perceiving the world around us in 3D from image sequences acquired
from a moving camera is a fundamental task in fields such as computer
vision, robotics, human-computer and human-robot
interaction. Visual SLAM (Simultaneous Localisation and
Mapping) systems have focused, for decades now, on jointly solving the
tasks of tracking the position of a camera as it explores unknown
locations and creating a 3D map of the environment. Their real-time
capability has turned SLAM methods into the cornerstone of
ambitious applications such as autonomous driving, robot navigation
and also augmented/virtual reality. Research in Visual SLAM
has progressed at a fast pace, moving from early works that
reconstructed sparse maps with just a few tens or hundreds of features
using filtering techniques~\cite{davison2003real}, to parallel
tracking and mapping approaches that could take advantage of
computationally expensive batch optimisation techniques for the
mapping thread to produce accurate maps with thousands of landmarks
% while maintaining a real-time camera tracking thread
~\cite{klein2007parallel,mur2015orb}, to contemporary methods that
allow instead to reconstruct completely \emph{dense} maps of the
environment~\cite{kinectfusion_2011,elasticfusion,newcombe2011dtam}.
The impact on \emph{augmented reality} of this progression
towards dense and robust real-time mapping has been
immense with many SLAM enabled \emph{augmented reality} applications
making their way into consumer products and mobile phone apps.

Despite these advances, there are still two areas in which SLAM
methods and their application to augmented reality are still very much
in their infancy.
\begin{itemize}
\item[\emph{(a)}] Most SLAM methods rely on the assumption that the
  environment is mostly static and moving objects are, at best,
  detected as outliers and ignored. Although some first steps have been taken
  towards non-rigid and dynamic scene
  reconstruction, with exciting results in reconstruction of a single
  non-rigid
  object~\cite{dynamicfusion,templatefusion,fusion4d,innmann2016volumedeform}
  or multiple moving rigid objects~\cite{cofusion}, designing an
  accurate and robust SLAM system that can deal with arbitrary dynamic
  and non-rigid scenes remains an open challenge.

\item[\emph{(b)}] The output provided by the majority of SLAM systems
  is a purely geometric map of the environment. The addition of
  semantic information is relatively
  recent~\cite{castle2007towards,civera2011towards,semanticfusion,cnn_slam_2017_cvpr,salas2013slam++}
  and is mostly limited to the recognition of a small number of
  known object instances for which a 3D model is available in
  advance~\cite{castle2007towards,civera2011towards,salas2013slam++,tateno2016icra}
  or to classify each 3D map point into a fixed set of semantic
  categories without differentiating  object
  instances~\cite{semanticfusion,cnn_slam_2017_cvpr}.
\end{itemize}

\noindent{\bf Contribution:} the novelty of our approach is to make
advances towards addressing both of these limitations within the same
system.
\methodtitle{} is a real-time capable SLAM system that can
represent scenes at the level of objects. It can recognise, detect,
track and reconstruct multiple moving rigid objects while precisely
segmenting each instance and assigning it a semantic label. We take
advantage of combining the outputs of: \emph{(i)}
Mask-RCNN~\cite{maskrcnn}, a powerful image-based instance level
segmentation algorithm that can predict object category labels for
$80$ object classes, and \emph{(ii)} a geometry-based segmentation
algorithm, that generates an object edge map from depth and surface
normal cues; to increase the accuracy of the object boundaries in the
object masks.

Our dynamic SLAM framework takes these accurate object masks as input
to track and fuse multiple moving objects (as well as the
static background) while propagating the semantic image labels into
temporally-consistent 3D map labels. The main advantage of using
instance-aware semantic segmentation over standard pixel-level
semantic segmentation (such as most previous semantic SLAM
systems~\cite{castle2007towards,civera2011towards,salas2013slam++,tateno2016icra,semanticfusion,cnn_slam_2017_cvpr})
is that it provides accurate object masks and the ability to segment
different object instances that belong to the same object category
instead of treating them as a single blob.

The additional advantage of \methodtitle{} over previous semantic SLAM
systems~\cite{castle2007towards,civera2011towards,salas2013slam++,tateno2016icra,semanticfusion,cnn_slam_2017_cvpr}
is that it does not require the scene to be static and so can
detect, track and map multiple independently moving
objects. Maintaining an internal 3D representation of moving objects
(instead of treating them as outliers) substantially improves the
overall SLAM system by providing a richer map that includes not just
the background but also the detailed geometry of the moving objects,
and by improving object and camera pose prediction and estimation.

On the other hand, the advantage of \methodtitle{} over previous
dynamic SLAM systems~\cite{cofusion,barsan2017simultaneous} is that it
enhances the dynamic map with semantic information from a large number
of object classes in real time. Not only can it detect individual
objects (thanks to the use of Mask-RCNN~\cite{maskrcnn}) and assign
semantic labels to their corresponding 3D map points, but it can also
accurately segment each individual object
instance. Table~\ref{tab:related_work} summarises our contributions in
the context of other \emph{real-time} semantic SLAM and dynamic SLAM
systems.

The result is a versatile system that can represent a
dynamic scene at the level of objects and their semantic labels, which
has numerous applications in areas such as robotics and augmented
reality. We demonstrate how the labels of objects can be used for
different purposes. For instance, we show that often, being able to
detect and segment \emph{people} allows us to be aware of their
presence, ignore those pixels and focus instead on the objects that
they are manipulating. We show how this can be useful in object
manipulation tasks, as it can improve object tracking even when objects
are moved and occluded by a human hand.

%%%%%%%%%%%%%%%%%%%%%%%%%%%%%%%%%%%%%%%%%%%%%%%%%%%%%%%%%%%%%%%%%%%%%%%%%%%%%%%%%%%%%%%%%%%%%%%%%%%%
\section{RELATED WORK}

% General related work here
The field of Visual SLAM has a long history of offering
solutions to the problem of jointly tracking the pose of a moving
camera (see~\cite{vslam_survey_2015} for a recent survey) while
reconstructing a map of the environment. The advent of inexpensive,
consumer-grade RGB-D cameras -- such as the Microsoft Kinect --
stimulated further research, and enabled the leap to \emph{dense}
real-time
methods~\cite{kinectfusion_2011,keller_2013_3DV,dense_rgbdslam_2013}.

\noindent{\bf Dense RGB-D SLAM:} Resulting methods are capable of
accurately mapping indoor environments and gained popularity in
augmented reality and robotics. KinectFusion~\cite{kinectfusion_2011}
proved that a truncated signed distance function (TSDF) based map
representation can achieve fast and robust mapping and tracking in
small environments. Subsequent
work~\cite{kintinuous,kfmoving_2012_bmvc} showed that the same
principles are applicable to large scale environments by choosing
appropriate data structures.

Surface elements (surfels) have a long history in computer
graphics~\cite{surfel_2000_siggraph} and have found many applications in
computer
vision~\cite{surfel_carceroni_2002_IJCV,hand_scanning_2009_ICCV}. More
recently, surfel-based map representations were also
introduced~\cite{rgbdmapping_2012_IJRR,keller_2013_3DV} to the domain
of RGBD-SLAM. A map of surfels is similar to a point cloud with the
difference that each element encodes local surface properties --
typically a radius and normal -- in addition to its
location. In contrast to a TSDF-based map, surfel clouds are naturally
memory efficient and avoid the overhead due to switching
representations between mapping and tracking that is typical of
TSDF-based fusion methods. Whelan et al.~\cite{elasticfusion} presented
a surfel-based RGBD-SLAM system for large environments with local and
global loop-closure.

\noindent{\bf Scene segmentation:} The computer
graphics~\cite{mesh_segmentation_bench_2009} and
vision~\cite{rgbdsegmentation_2011_ICRA, karpathy2013icra,
  patchvolumes_2013_3dv, tateno2016icra, semanticfusion} communities
have devoted substantial effort to object and scene segmentation.
Segmented data can broaden the functionality of visual tracking and mapping systems, for
instance, by enabling robots to detect objects. Some methods have
proposed to segment RGBD data based on geometric properties of
surface normals~\cite{rgbdsegmentation_2011_ICRA, karpathy2013icra,
  finman2014, tateno2015iros}, mainly by assuming that objects are
convex. While the clear strength of geometry-based segmentation
systems is that they produce accurate object boundaries, their
weakness is that they typically result in over-segmentations and they
do not convey any semantic information.

\noindent{\bf Semantic scene segmentation:} Another line of
work~\cite{pointcloud_labelling_2005_CVPR,
  pointcloud_labelling_2010_BMVC, pointcloud_labelling_2011_NIPS} aims
at segmenting 3D scenes semantically, using Markov Random Fields
(MRFs). These methods require labelled 3D data, however, which in
contrast to labelled 2D image data is not readily available. This is
exemplified by the fact that all three works involved manual
annotation of training data. Datasets containing isolated RGBD frames,
such as NYUv2~\cite{nyu2_2012_ECCV}, are not applicable here and it
requires significant effort to build consistent reconstructed datasets
for segmentation, as recently shown by Dai et
al.~\cite{scannet_2017_cvpr}.

\noindent{\bf Semantic SLAM:} Motivated by the success of
convolutional neural networks\cite{fasterrcnn, sharpmask, maskrcnn},
Tateno et al.~ \cite{cnn_slam_2017_cvpr} and McCormac et
al.~\cite{semanticfusion} integrate deep neural networks in real-time
SLAM systems. As inference is solely based on 2D information, the need
for 3D annotated data is circumvented. The resulting systems offer
strategies to fuse labelled image data into segmented 3D maps. Earlier
work by Hermans et al.~\cite{densesemantic_2014_icra} implements a
similar scheme, using a randomised decision forest
classifier. However, since the systems are not considering object
instances, tracking multiple models independently is unattainable.

\noindent{\bf Dynamic SLAM:} There are two main scenarios in dynamic
SLAM: non-rigid surface reconstruction and multibody formulations for
independently moving rigid objects. In the first case, a deformable
world is assumed~\cite{dynamicfusion, fusion4d, templatefusion} and
as-rigid-as-possible registration is performed, while in the second,
rigid object instances are
identified~\cite{tateno2016icra,salas2013slam++} and tracked
sparsely~\cite{slammot_2007_IJRR, coslam_2013_PAMI} or densely
~\cite{cofusion}. Both categories use template- or descriptor-based
formulations~\cite{tateno2016icra,templatefusion,salas2013slam++},
which require pre-observing objects of interest, and template-free
methods.
In the case when the dynamic parts of the scene are not of interest,
it is valuable to recognise them as outliers to avoid errors in the
optimisation back-end. Methods for the explicit detection of dynamic
regions for static fusion were proposed by Jaimez et
al.~\cite{vosf_2017_ICRA} and Scona et
al.~\cite{staticfusion_2018_ICRA}.

Table~\ref{tab:related_work} provides an overview of related {\bf real-time}
capable methods comparing them under five important properties.

Only two dynamic SLAM system, to the best of our knowledge,
have previously attempted incorporates semantic
knowledge, but both fall short of the functionality of
\methodtitle{}. Co-Fusion~\cite{cofusion} demonstrated the ability to
track, segment and reconstruct objects based on their semantic labels,
but the overall system was not real-time capable and limited
functionality was shown. DynSLAM~\cite{barsan2017simultaneous}
developed a mapping system for autonomous driving applications capable
of separately reconstructing both the static environment and the
moving vehicles. However, the overall system was not real-time (this
is the reason it does not appear in Table~\ref{tab:related_work}) and
\emph{vehicle} was the only dynamic object-class it reconstructed,
so its functionality was limited to road scenes.

% groups

\begin{table}
  \center
  \setlength\tabcolsep{4pt}
    \begin{tabular}{c||l|l|l||l|l}
    Method       & \specialcell[b]{Model-\\free}  & \specialcell[b]{Scene\\Segmen-\\tation}  & \specialcell[b]{Semantics} & \specialcell[b]{Multiple\\moving\\objects} & \specialcell[b]{Non-\\Rigid}  \\ \hline \hline
    \specialcell[b]{Static-\\Fusion\cite{staticfusion_2018_ICRA}} & \checkmark                                                 & \checkmark                               & ~                      & ~      & ~                             \\ \hline

    \specialcell[b]{2.5D is not\\enough\cite{tateno2016icra}}     &  ~                                                                     & \checkmark                               & \checkmark            & ~       & ~                             \\ \hline
    \specialcell[b]{Slam++\\\cite{salas2013slam++}}     &                                                                   & \checkmark                               & \checkmark            & ~       & ~                             \\ \hline

    \specialcell[b]{CNN-\\SLAM\cite{cnn_slam_2017_cvpr}}          & \checkmark                                                             & \checkmark                               & \checkmark            & ~       & ~                             \\ \hline
    \specialcell[b]{Semantic-\\Fusion\cite{semanticfusion}}       & \checkmark                                                                 & \checkmark                               & \checkmark         & ~          & ~                              \\ \hline\hline
    \specialcell[b]{Non-Rigid\\RGBD\cite{templatefusion}}         & ~                                 &                   ~                                                                & ~    & ~                        & \checkmark                    \\ \hline
    \specialcell[b]{Dynamic-\\Fusion\cite{dynamicfusion}}         & \checkmark                                                & ~                                        & ~    & ~                        & \checkmark                    \\ \hline
    \specialcell[b]{Fusion4D\\\cite{fusion4d}}                    & \checkmark                                              & ~                                        & ~      & ~                      & \checkmark                    \\ \hline
    \specialcell[b]{Co-\\Fusion\cite{cofusion}}                   & \checkmark                                                     & \checkmark                               & ~       & \checkmark                    & ~                             \\ \hline
    \textbf{\specialcell[b]{Mask-\\Fusion}}                       & \checkmark                                                       & \checkmark                               & \checkmark & \checkmark                 & ~                             \\
    \end{tabular}
    \caption{Comparison of the properties of \methodtitle{} with respect
      to other {\bf real-time} SLAM systems. In contrast to previous
      semantic SLAM
      systems\cite{tateno2016icra,salas2013slam++,cnn_slam_2017_cvpr,semanticfusion},
      \methodtitle{} is both dynamic (it reconstructs objects even
      when their motion is different from the camera) and segments
      object instances. Unlike dense non-rigid reconstruction
      systems~\cite{templatefusion,dynamicfusion,fusion4d}, it can
      reconstruct the entire scene and adds semantic labels to
      different objects. Note that while Co-Fusion~\cite{cofusion} could use
      semantic cues to segment the scene, in that case the system
      was not real-time -- only the non-semantic version of Co-Fusion was real-time capable.}
    \label{tab:related_work}
\end{table}

%%%%%%%%%%%%%%%%%%%%%%%%%%%%%%%%%%%%%%%%%%%%%%%%%%%%%%%%%%%%%%%%%%%%%%%%%%%%%%%%%%%%%%%%%%%%%%%%%%%%
\section{System Overview}\label{sec:overview}

\begin{figure}[t]
 \subfloat[Timing of asynchronous components: In this timeline, frame
   \textbf{S} and frame \textbf{M} are highlighted with thick borders, as
   the SLAM and masking threads are working on them respectively.
   \textbf{C}, the current frame (tail of queue $Q_f$) is shown
   in blue, the head of the queue is shaded in green, and frames with
   available object masks are marked orange.]
          {\includegraphics[width=\linewidth]{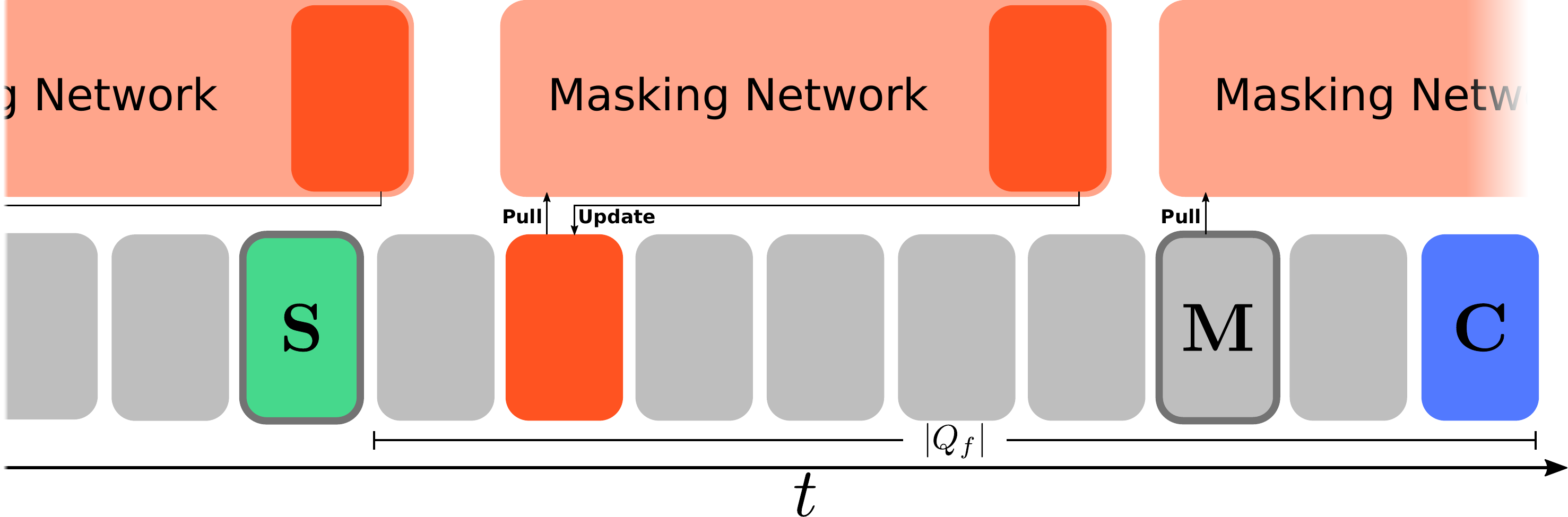}\label{fig:queue}}
          \\ \subfloat[Dataflow in \methodtitle{}: Camera frames are
            added to a fixed length queue $Q_f$. The SLAM system
            (green) operates on its head. The semantic masking DNN
            pulls input frames from the tail, and updates frames back
            to the queue as soon as results (semantic masks) are
            available.]{\includegraphics[width=\linewidth]{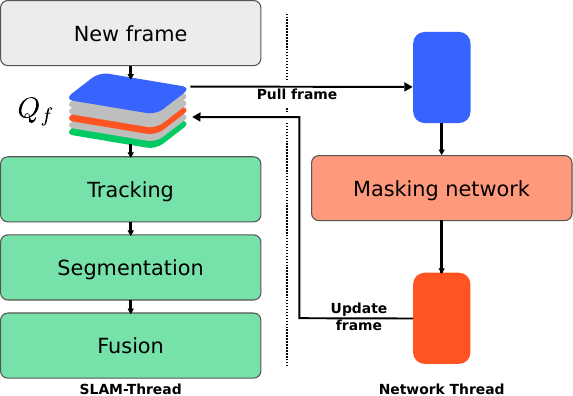}}
\caption{High-level overview of the SLAM back-end and masking network,
  and their interaction.}
\label{fig:flow}
\end{figure}

\methodtitle{} enables real-time dense dynamic RGBD SLAM at the level of
objects. In essence, \methodtitle{} is a multi-model SLAM system that
maintains a 3D representation for each object that it recognises in
the scene (in addition to the background model). Each model is tracked
and fused independently. Figure~\ref{fig:flow} illustrates its
frame-to-frame operation. Each time a new frame is acquired by the camera, the following steps
are performed:\\
\noindent{\bf Tracking:} The 3D geometry of each object is represented
as a set of surfels. The six degree of freedom pose of each model is
tracked by minimizing an energy that combines a geometric iterative
closest point (ICP) error with a photometric cost based on brightness
constancy between corresponding points in the current frame and the
stored 3D model, aligned with the pose in the previous frame. In order
to lower computational demand and increase robustness, only non-static
objects are tracked separately. Two different strategies were tested
to decide whether an object is static or not: one based on motion
inconsistency, similar to~\cite{cofusion}, and another that treats
objects which are being touched by a person as dynamic.\\
\noindent{\bf Segmentation:} \methodtitle{} combines two types of cues
for segmentation: semantic and geometric cues. Mask-RCNN~\cite{maskrcnn} is
used to provide object masks with semantic labels. While this
algorithm is impressive and provides good object masks, it suffers from
two drawbacks. First, the algorithm does not run in real time and can
only operate at a maximum of 5 Hz. Second, the object boundaries
are not perfect -- they tend to leak into the background. To overcome
both of these limitations, we run a geometric segmentation
algorithm, based on an analysis of depth discontinuities and surface
normals. In contrast to the semantic instance segmentation, the geometric
segmentation runs in real time and produces very accurate object
boundaries (see Figures
\ref{fig:frame}(d) and (e) for an example visualisation of the geometric edge
map and the geometric components returned by the algorithm). On the
negative side, geometry-based segmentation tends to oversegment
objects. The combination of these two segmentation strategies  -- geometric
segmentation on a per-frame basis and semantic segmentation as often
as possible -- provides
the best of both worlds, allowing us to (1) run an overall system in real time
(geometric segmentation is used for frames without semantic object
masks, while the combination of both is used for frames with object
masks) and (2) obtain semantic object masks with improved object
boundaries, thanks to the geometric segmentation.
\\
\noindent{\bf Fusion:} The geometry of each object is fused over time
by using the object labels to associate surfels with the correct
model. Our fusion follows the same strategy as~\cite{keller_2013_3DV,
  elasticfusion}. \\

The rest of the paper is organised as follows. We first describe the
principles of our dynamic RGBD-SLAM method in Section~\ref{sec:slam};
further details regarding the integration of the semantic and
geometric segmentation results are provided in
Section~\ref{sec:segmentation}.  A quantitative and qualitative
evaluation of the proposed approach is presented in
Section~\ref{sec:evaluation}.

%%%%%%%%%%%%%%%%%%%%%%%%%%%%%%%%%%%%%%%%%%%%%%%%%%%%%%%%%%%%%%%%%%%%%%%%%%%%%%%%%%%%%%%%%%%%%%%%%%%%
\section{MULTI-OBJECT SLAM}\label{sec:slam}

\methodtitle{} maintains a set of independent 3D models, $\mathcal{M}_m \;
\forall m \in \{0..N\}$, for each of the $N$ objects recognised in the
scene and a further model for the background. We adopt the surfel
representation popularised by~\cite{keller_2013_3DV, elasticfusion},
where a model ${\mathcal{M}_m}$ is represented by a cloud of surfels ${\mathcal{M}_m^s
  \in (\mathbf{p}\in\mathbb{R}^3, \mathbf{n}\in\mathbb{R}^3,
  \mathbf{c}\in\mathbb{N}^3, \mathbf{w}\in\mathbb{R},
  \mathbf{r}\in\mathbb{R}, \mathbf{t}\in \mathbb{R}^2)} \; \forall s <
\vert \mathcal{M}_m \vert$, which are tuples of position, normal,
colour, weight, radius and two timestamps. Additionally, models are
associated with a class ID $c_m \in \{0..80\}$ and an object-label
$l_m=m \; \forall m \in \{0..N\}$. Finally, for each time instance $t$, an
\textit{is static} indicator $s_{tm} \in 0,1$ and a rigid pose
$\mathbf{R}_{tm} \in \mathbb{SO}_3$, $\mathbf{t}_{tm} \in
\mathbb{R}^3$ is stored.

\subsection{Tracking}

Assuming that a good estimate exists for the pose of model
$\mathcal{M}_m$ at time $t-1$, the pose at time $t$ is inferred by
aligning the current depth-map $\mathcal{D}_t$ and intensity-map
$\mathcal{I}_t$ with the projection $\mathcal{D}^a_{t-1},
\mathcal{I}^a_{t-1}$ of $\mathcal{M}_m$, which is generated by
rendering its surfels using the OpenGL pipeline. Here, $\mathcal{D}_t$
and $\mathcal{I}_t$ are mappings from image-coordinates $\Omega \subset
\mathbb{N}^2$ to depth $\mathcal{D}_t: \Omega\rightarrow\mathbb{R}$ and
grey-scale $\mathcal{I}_t: \Omega\rightarrow\mathbb{N}$,
respectively. $\mathcal{I}_t$ is derived by weighting RGB channels as
follows: $r,g,b \mapsto 0.299r + 0.587g + 0.114b$.

The alignment is performed by minimising a joint geometric and
photometric error function~\cite{cofusion, elasticfusion}:
\begin{equation}\label{eqn:energy_tracking}
E_m = \min_{{\mathbf{\xi}_m}}\,\, (E^{icp}_m + \lambda E^{rgb}_m),
\end{equation}

where $E^{icp}_m$ and $E^{rgb}_m$ are the geometric and photometric error
terms respectively and  $\mathbf{\xi}_m$ is the unknown
rigid transformation, expressed in a minimal 6D Lie algebra
representation $\mathfrak{se}_3$, which is subject to optimisation.

The first term in equation~(\ref{eqn:energy_tracking}) is a sum of
projective ICP residuals. Given a vertex $\mathbf{v}_t^i$, which is
the back-projection of the $i$-th vertex in $\mathcal{D}_t$; and
$\mathbf{v}^i$ and $\mathbf{n}^i$, the corresponding vertex and normal
in $\mathcal{D}^a_{t-1}$ (the geometry expressed in the camera
coordinate frame at time $t-1$), $E^{icp}_m$ is written as:
\begin{equation}\label{eqn:icp_error}
E^{icp}_m = \sum_i
\left( (\mathbf{v}^i -
\exp(\mathbf{\xi}_m) \mathbf{v}_t^i) \cdot \mathbf{n}^i \right)^2
\end{equation}

The photometric term, on the other hand, is a sum of photo-consistency
residuals between $\mathcal{I}_t$ and $\mathcal{I}^a _{t-1}$,
and reads as follows:
\begin{equation}\label{eqn:photometric_error}
E^{rgb}_m = \sum_{\mathbf{u} \in \Omega}
\left(\mathcal{I}_t(\mathbf{u}) -
\mathcal{I}^a _{t-1}(\pi(\exp(\mathbf{\xi}_m) \pi^{-1}(\mathbf{u},\mathcal{D}_t)))\right)^2
\end{equation}

Here, $\pi$ performs a perspective projection $\pi: \mathbb{R}^3
\rightarrow \mathbb{R}^2$, whereas $\pi^{-1}$ back-projects from a
depth map with 2D coordinate. To optimise this non-linear
least-squares cost we use a Gauss-Newton solver with a four level
coarse-to-fine pyramid scheme. The CUDA accelerated implementation of
the solver builds on the open source code releases
of~\cite{elasticfusion} and ~\cite{cofusion}.

\subsection{Fusion}

Given $\mathbf{R}_{tm}$ and $\mathbf{t}_{tm}$, surfels for each model $\mathcal{M}_m$ are updated by
performing a projective data association with the current RGBD frame. This step is inspired
by~\cite{keller_2013_3DV} but a stencilling based on the segmentation discussed in
Section~\ref{sec:segmentation} is used to adhere to object boundaries. As a result, each newly
created surfel is part of exactly one model. Further, we introduce a confidence penalty for surfels
outside the stencil, which is required due to imperfect segmentations.

%%%%%%%%%%%%%%%%%%%%%%%%%%%%%%%%%%%%%%%%%%%%%%%%%%%%%%%%%%%%%%%%%%%%%%%%%%%%%%%%%%%%%%%%%%%%%%%%%%%%
\section{SEGMENTATION}\label{sec:segmentation}

\methodtitle{} reconstructs and tracks multiple objects
simultaneously, maintaining separate models. As a consequence, new
data has to be associated with the correct model before fusion is
performed.  Inspired by Co-Fusion~\cite{cofusion}, instead of
associating data in 3D, segmentation is carried out in 2D and
model-to-segment correspondences are established. Given these
correspondences, new frames are masked and only subsets of the data
are fused with existing models.  Masking is based on the semantic
instance segmentation labels proposed by a DNN~\cite{maskrcnn}, in
conjunction with geometric segmentation, which improves the quality
of object boundaries. Our semantic segmentation pipeline provides
masks at 30Hz or more.

The design of the pipeline is based on the following observations:
\textit{(i)} Current semantic segmentation methods are good at
detecting objects, but tend to provide imperfect object
boundaries. \textit{(ii)} The current state-of-the-art approach,
Mask-RCNN~\cite{maskrcnn}, cannot be executed at frame
rate. \textit{(iii)} The information contained in RGBD frames enables
fast over-segmentation of the image, for instance by assuming object
convexity.

The second observation directly implies that to achieve overall
real-time performance our system must execute instance level semantic
segmentation in a parallel thread concurrently to the tracking and
fusion threads. However, executing two programs at different
frequencies concurrently requires a synchronisation strategy. We
buffer new frames in a queue $Q_f$ and refer the SLAM system to the
head of the queue, while the semantic segmentation operates on the
back of the queue, as illustrated in Figure~\ref{fig:queue}. This way,
the execution of the SLAM pipeline is delayed by the worst-case
processing time of the semantic segmentation. In our experiments we
picked a queue length of 12 frames, which involves a delay of
approx. 400ms. Whether this delay can be neglected or not, depends on
the use-case of the system. Even though a latency exists, the system
runs at a frame-rate of 30fps.  Furthermore, a semantic segmentation
is not available for most frames due to the lower execution frequency
of the masking component, yet each frame requires a labelling in order
to fuse new data. This issue is solved by associating regions of
mask-less frames with existing models only, as discussed in
Section~\ref{sec:merge}.

To compensate for inexact boundaries, as mentioned in observation~\textit{1}, we make use of
observation~\textit{3} and map components from a geometric over-segmentation to semantic masks. This
results in improved masks, due to higher-quality boundaries of the geometric segmentation.

\begin{figure}
 \centering
 \subfloat[RGB input]{\includegraphics[width=.49\linewidth]{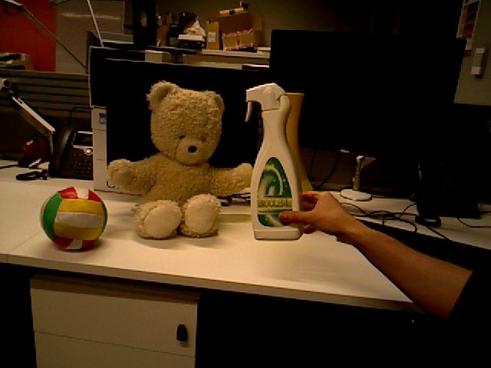}\label{fig:frame:color}}
 \subfloat[Depth map]{\includegraphics[width=.49\linewidth]{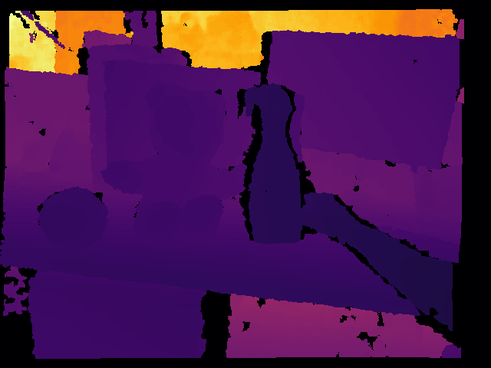}\label{fig:frame:depth}}
 \\
 \subfloat[Semantic masks]{\includegraphics[width=.49\linewidth]{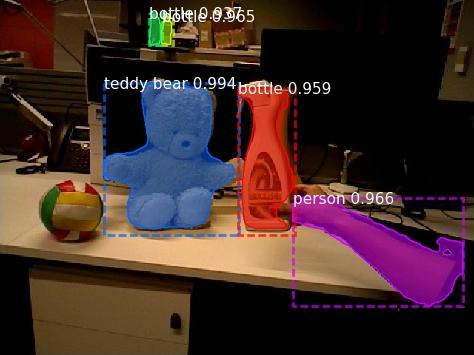}\label{fig:frame:mask}}
 \subfloat[Geometric edges]{\includegraphics[width=.49\linewidth]{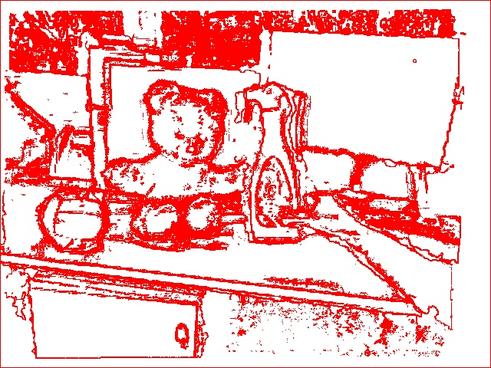}\label{fig:frame:geom}}
 \\
 \subfloat[Components]{\includegraphics[width=.32\linewidth]{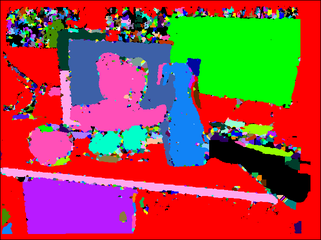}\label{fig:frame:components}}
 \subfloat[Projected labels]{\includegraphics[width=.32\linewidth]{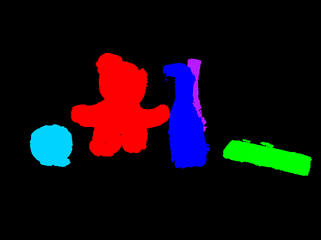}\label{fig:frame:projected}}
 \subfloat[Final segmentation]{\includegraphics[width=.32\linewidth]{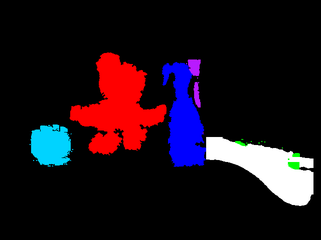}\label{fig:frame:result}}
 \\
 \subfloat[Reconstructed objects]{\includegraphics[width=.99\linewidth]{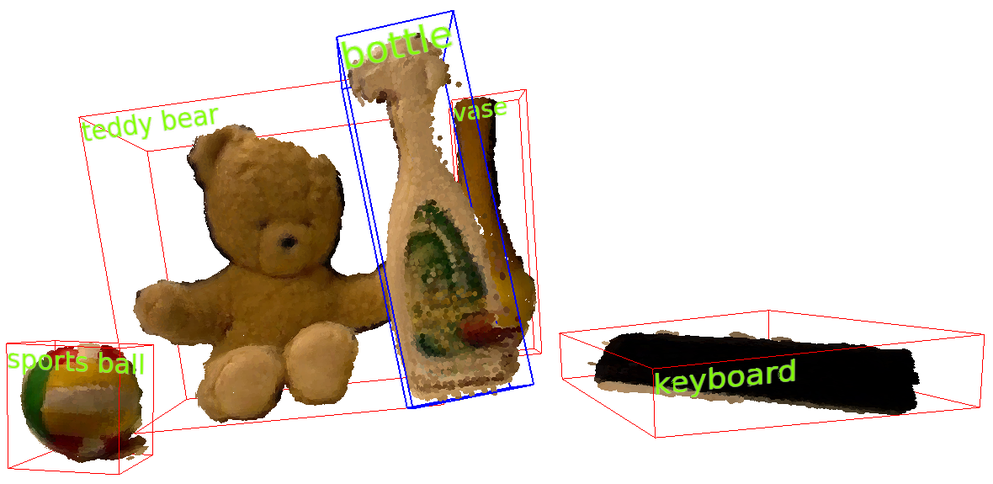}\label{fig:frame:rec}}
 % \subfloat[Reconstructed normals]{\includegraphics[width=.49\linewidth]{segmentation/frame_900_geom}\label{fig:frame:normals}}
 % \subfloat[Geometric edges]{\includegraphics[width=.48\linewidth]{frame_geom}\label{fig:frame:geom}}
 \caption{Breakdown of the segmentation method. While~(a) and~(b) show an input RGBD frame, (c)-(g) visualise the output of different stages.}
 \label{fig:frame}
\end{figure}

\begin{figure}
  \subfloat[Segment of interest]{\includegraphics[width=.48\linewidth]{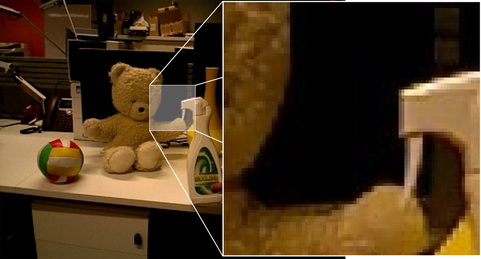}}~
    \subfloat[Semantic only]{\includegraphics[width=.26\linewidth]{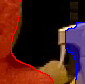}}~
  \subfloat[With geometric]{\includegraphics[width=.26\linewidth]{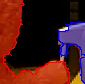}}
  \caption{Comparison of boundaries produced by semantic labelling only and by merged semantic and geometric labelling. While the semantic segmentation is smooth, it lacks important details.}
  \label{fig:segdiff}
\end{figure}

\subsection{SEMANTIC INSTANCE SEGMENTATION}

A variety~\cite{Li_2017_CVPR, sharpmask, maskrcnn} of recently proposed neural network architectures
are tackling the problem of instance-level object segmentation. They outperform traditional methods
and are capable of handling a large set of object classes. Of these methods,
Mask-RCNN~\cite{maskrcnn} is especially compelling, as it provides superior segmentation quality at
a relatively high frame-rate of 5Hz. The semantic segmentation pipeline of \methodtitle{} is based
on Mask-RCNN\footnote{We are using the Matterport~\cite{maskrcnnimpl} implementation of Mask-RCNN.},
which maps RGB frames to a set of object masks $\mathcal{L}_{tn}^s: \Omega\rightarrow \{0,1\}$,
bounding boxes $\mathbf{b}_{tn} \in \mathbb{N}^4$ and class IDs $c_{tn} \in \{0..80\}$, for all $n \in
\{1..N_t^s\}$ of the $N_t^s$ instances detected in the frame at time $t$.
% $\mathcal{C}_t: \Omega\rightarrow\mathbb{N}^3$

Mask-RCNN achieves this by extending the Faster-RCNN~\cite{fasterrcnn} architecture. Faster-RCNN is
a two-stage approach that proposes regions of interest first and then predicts an object class and
bounding box per region and in parallel. He et al. added a third branch to the second stage, which
generates masks independently of class IDs and bounding boxes.
Both stages rely on a feature map, which is extracted by a ResNet\cite{he2016deep}-based backbone network, and apply
convolutional layers for inference.

Figure~\ref{fig:frame:mask} visualises the output of Mask-RCNN. Note that instances of the same
class are highlighted with different colours, and also that masks are not perfectly aligned with
object boundaries.

\subsection{GEOMETRIC SEGMENTATION}

Assuming that objects -- especially man-made objects -- are largely
convex, it is possible to build fast segmentation methods that place
edges in concave areas and depth discontinuities. In practice, such
methods tend to oversegment data, due to the simplified
premise. Moosmann et al.~\cite{moosmann2009} successfully segment 3D
laser data based on this assumption. The same principle is also used
by other authors to segment objects in RGBD
frames~\cite{rgbdsegmentation2012, karpathy2013icra, finman2014,
  tateno2015iros, stein2014cvpr}.

Our geometric segmentation method follows this approach and, similarly to \cite{tateno2015iros},
generates an edginess-map based on a depth discontinuity term $\phi_d$ and concavity term $\phi_c$.
Specifically, a pixel is defined as an edge pixel if $\phi_d + \hat{\lambda} \phi_c > \tau$, where
$\tau$ is a threshold and $\hat{\lambda}$ a relative weight. Given a local neighbourhood
$\mathcal{N}$, $\phi_d$ and $\phi_c$ are computed as follows:

\begin{equation}
 \phi_d = \underset{i \in \mathcal{N}}{\max}
 \vert (\mathbf{v}_i - \mathbf{v}) \cdot \mathbf{n} \vert
 \label{eq:phid}
\end{equation}

\begin{equation}
 \phi_c = \underset{i \in \mathcal{N}}{\max}
 \begin{cases}
  0                 & \quad \text{if } (\mathbf{v}_i - \mathbf{v}) \cdot \mathbf{n} < 0 \\
  1 - (\mathbf{n}_i \cdot \mathbf{n}) & \quad \text{else}
 \end{cases}
 \label{eq:phic}
\end{equation}

Here, $\mathbf{v}$ and $\mathbf{v}_i$ indicate vertex positions, while $\mathbf{n}$ and
$\mathbf{n}_i$ represent normals, obtained by back-projecting $\mathcal{D}_t$. Since $\phi_d +
\hat{\lambda} \phi_c$ depends on a local neighbourhood only, the edginess of a pixel can be
evaluated quickly on a GPU. Figure~\ref{fig:frame:geom} shows the edge
map for a frame that was
captured with an Asus Xtion RGBD-camera. Edge maps are converted to a geometric labelling
$\mathcal{L}_{t}^g: \Omega\rightarrow \{0..N_t^g\}$, where $N_t^g$ is the number of extracted
components excluding the background, by running an out-of-the-box connected components algorithm, as
illustrated in Figure~\ref{fig:frame:components}.

\subsection{MERGED SEGMENTATION}\label{sec:merge} % FUSING?

\begin{figure}
 \includegraphics[width=\linewidth]{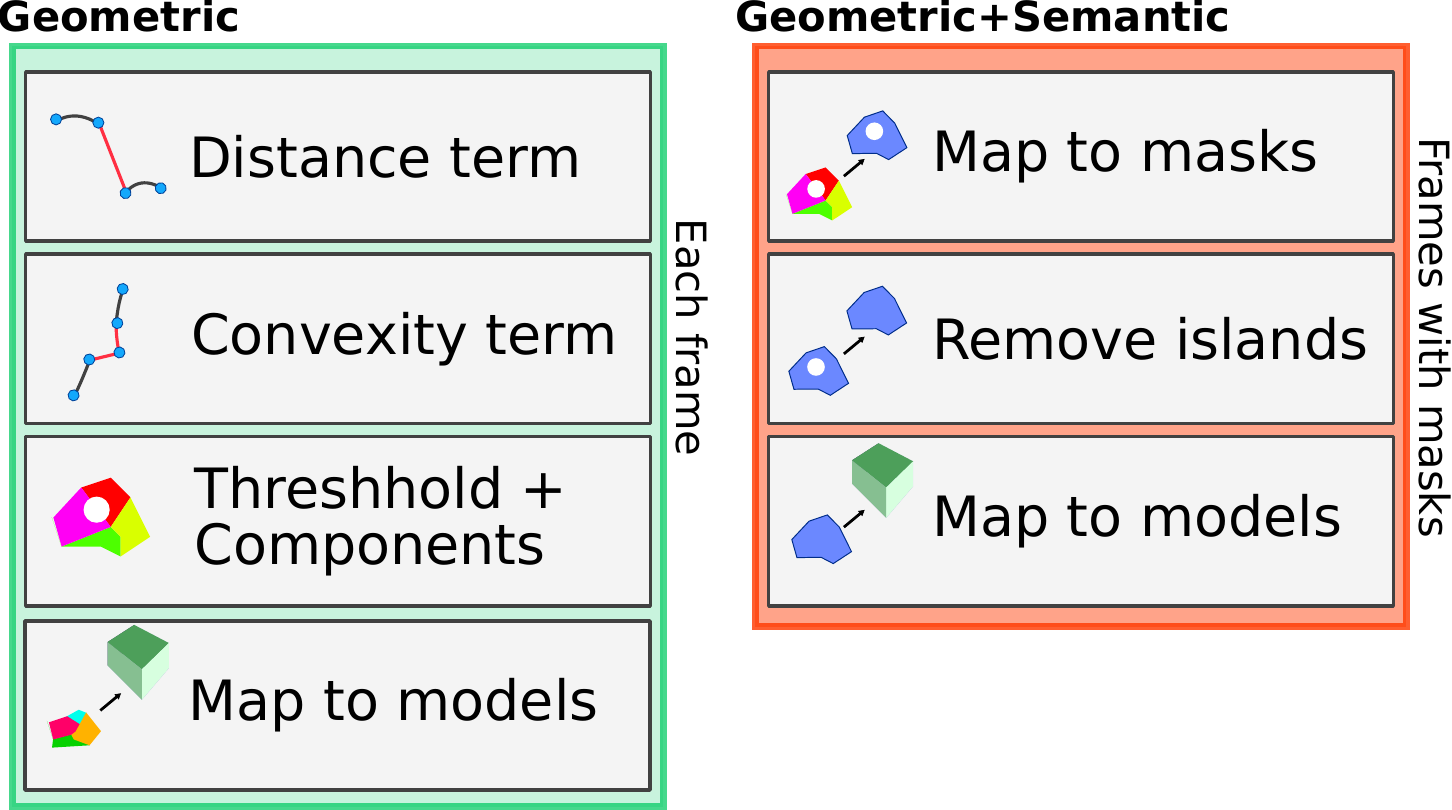}
 \caption{Overview of performed segmentation steps. A geometric segmentation is performed for each frame and resulting components are mapped to masks if available, which in turn are mapped to existing models. Components that are not mapped to masks are directly associated with an object, if possible.}
 \label{fig:segmentation}
\end{figure}

For each frame that is processed by the SLAM system, the pipeline illustrated in
Figure~\ref{fig:segmentation} is executed. While the geometric segmentation, shown on the
left-hand-side, is performed for all frames, geometric labels are mapped to semantic masks only if
these are available. In the absence of semantic masks, geometric labels are associated with existing
models directly and the following steps are skipped:

\subsubsection{Mapping geometric labels to masks}
%\textit{Mapping geometric labels to masks.}
After over-segmenting input frames geometrically, the resulting components $C_{ti} \; \forall i \in
\{1..N_t^g\}$ are mapped to masks $\mathcal{L}_{tn}^s$ by identifying the one with maximal overlap.
Only if this overlap is greater than a threshold -- in our experiments $65\% \cdot |C_{ti}|$, where
$|C_{ti}|$ denotes the number of pixels belonging to component $C_{ti}$ -- a mapping is assigned.
Note that multiple components can be mapped to the same mask, but no more than a single mask is
linked to a component. An updated labelling $\mathcal{L}_{t}^c: \Omega\rightarrow
{1..N_t^s}$ is computed, which replaces component with mask IDs, if an assignment was made.

\subsubsection{Mapping masks to models} Next, a similar overlap between grouped components $C_{tj}$
in $\mathcal{L}_{t}^c$ and projected object labels $\mathcal{L}^a$, as shown in
Figure~\ref{fig:frame:projected}, is evaluated. Requiring that the camera and objects are tracked
correctly, $\mathcal{L}^a$ is generated by rendering all models using the OpenGL pipeline. Besides
testing an analogous threshold to before ($5\% \cdot |C_{tj}|$), it is verified that the
object class IDs of model and mask coincide. \\

Components that are not yet assigned to a model are now considered to be assigned directly. This is
necessary because Mask-RCNN can fail to recognise objects, and most frames are expected to not
exhibit any masks. Once again, an overlap of $65\% \cdot |C_i|$ between remaining components and
labels in $\mathcal{L}^a$ is evaluated.

The final segmentation $\mathcal{L}_{t}: \Omega\rightarrow \{0..N\}$ contains the object ids of the
models associated with relevant components. A special pre-defined value\footnote{We use the value
$255$, as we represent labels as unsigned bytes and assume a number of models less than that.} is
used to specify areas that ought to be ignored during fusion. This is especially useful to
explicitly prevent the reconstruction of certain object classes, such as the arm of the person in
Figure~\ref{fig:frame:result}, highlighted in white.

%%%%%%%%%%%%%%%%%%%%%%%%%%%%%%%%%%%%%%%%%%%%%%%%%%%%%%%%%%%%%%%%%%%%%%%%%%%%%%%%%%%%%%%%%%%%%%%%%%%%
\section{EVALUATION}\label{sec:evaluation}

Since the mapping and tracking components of \methodtitle{} are based on the work of \cite{cofusion,
elasticfusion}, we focus on the ability to tackle challenging problems that are not solvable by traditional SLAM
systems and refer the reader to the corresponding publications for additional details.

\subsection{Quantitative results}

\subsubsection{Trajectory estimation}

\begin{table}
\center
 \setlength\tabcolsep{4.45pt}
 \begin{tabular}{|l|l|l|l|l|l|l|}
  \hline
  Setting                                             	& Sequence          & VO-SF & EF            & CF    & SF  & MF  \\ \hline \hline
  \multirow{3}{*}{\specialcell[c]{Slightly\\dynamic}} 	& f3s\_static       & 2.9   & \textbf{0.9}  & 1.1   & 1.3  & 2.1  \\
                                                      	& f3s\_xyz          & 11.1  & \textbf{2.6}  & 2.7   & 4.0  & 3.1  \\
                                                      	& f3s\_halfsphere   & 18.0  & 13.8          & \textbf{3.6}   & 4.0  & 5.2  \\  \hline
  \multirow{3}{*}{\specialcell[c]{Highly\\dynamic}} 	  & f3w\_static       & 32.7  & 6.2           & 55.1  & \textbf{1.4}  & 3.5  \\
                                                      	& f3w\_xyz          & 87.4  & 21.6          & 69.6  & 12.7 & \textbf{10.4}  \\
                                                      	& f3w\_halfsphere   & 73.9  & 20.9          & 80.3\hphantom{0}  & 39.1 & \textbf{10.6}  \\ \hline
  \multicolumn{7}{c}{\rule{0pt}{9pt}(a) Comparison of AT-RMSEs (cm)} \vspace{0.3cm}
  \end{tabular}

  \begin{tabular}{|l|l|l|l|l|l|l|}
  \hline
  Setting                                             	& Sequence          & VO-SF & EF             & CF             & SF            & MF           \\ \hline \hline
  \multirow{3}{*}{\specialcell[c]{Slightly\\dynamic}} 	& f3s\_static       & 2.4   & \textbf{1.0}   & 1.1            & 1.1           & 1.7          \\
                                                      	& f3s\_xyz          & 5.7   & 2.8            & \textbf{2.7}   & 2.8           & 4.6          \\
                                                      	& f3s\_halfsphere   & 7.5   & 10.2           & \textbf{3.0}   & \textbf{3.0}  & 4.1          \\  \hline
  \multirow{3}{*}{\specialcell[c]{Highly\\dynamic}} 	  & f3w\_static       & 10.1  & 5.8            & 22.4           & \textbf{1.3}  & 3.9          \\
                                                      	& f3w\_xyz          & 27.7  & 21.4           & 32.9           & 12.1          & \textbf{9.7} \\
                                                      	& f3w\_halfsphere   & 33.5  & 16.3           & 40.0\hphantom{0}           & 20.7          & \textbf{9.3\hphantom{0}} \\ \hline
  \multicolumn{7}{c}{\rule{0pt}{9pt}(b) Comparison of translational RP-RMSEs (cm/s)}\vspace{0.3cm}
  \end{tabular}

  \begin{tabular}{|l|l|l|l|l|l|l|}
  \hline
  Setting                                             	& Sequence          & VO-SF & EF            & CF             & SF            & MF            \\ \hline \hline
  \multirow{3}{*}{\specialcell[c]{Slightly\\dynamic}} 	& f3s\_static       & 0.71  & \textbf{0.32}	& 0.44           & 0.43          & 0.54          \\
                                                      	& f3s\_xyz          & 1.44  & \textbf{0.77}	& 1.00           & 0.92          & 1.25          \\
                                                      	& f3s\_halfsphere   & 2.98  & 3.20	        & \textbf{1.92}  & 2.11          & 2.07          \\  \hline
  \multirow{3}{*}{\specialcell[c]{Highly\\dynamic}} 	  & f3w\_static       & 1.68  & 1.06	        & 4.01           & \textbf{0.38} & 0.76          \\
                                                      	& f3w\_xyz          & 5.11  & 4.31	        & 5.55           & 2.66          & \textbf{2.00} \\
                                    				  	        & f3w\_halfsphere   & 6.69  & 4.47	        & 13.02          & 5.04          & \textbf{3.35} \\ \hline
  \multicolumn{7}{c}{\rule{0pt}{9pt}(c) Comparison of rotational RP-RMSEs (deg/s)}
  \end{tabular}
  \caption{Quantitative comparison to other methods.}
  \label{tab:eval}
\end{table}

To objectively compare \methodtitle{} with other methods, we evaluate its performance on an
established RGBD benchmark dataset~\cite{tum_2012_iros}. This dataset offers sequences of colour and
depth frames and includes ground-truth camera poses to compare with. Measures commonly used for the
analysis of visual SLAM or visual odometry methods are the absolute trajectory error (ATE) and the
relative pose error (RPE). While the ATE evaluates the overall quality of a trajectory by summing
positional offsets of ground-truth and reconstructed locations, the RPE considers local motion
errors and therefore surrogates drift. To provide scene-length independent measures, both entities
are usually expressed as root-mean-square-error (RMSE). Since \methodtitle{} is designed to work in
dynamic environments, we chose according sequences from the dataset.

First, we estimate camera motion on scenes that involve rapid movement of persons. As our method --
as with the methods to which we compare -- is not capable of reconstructing deformable parts, we
exploit the contextual knowledge of \methodtitle{} to neglect data associated with persons.
Table~\ref{tab:eval} lists AT-RMSE and RP-RMSE measurements of five methods, including
\methodtitle{} (MF):
\begin{itemize}
  \item VO-SF~\cite{vosf_2017_ICRA}: A close to real-time method that computes piecewise-rigid scene
        flow to segment dynamic objects.
  \item ElasticFusion EF~\cite{elasticfusion}: A visual SLAM system that assumes a static
        environment.
  \item Co-Fusion (CF)~\cite{cofusion}: A visual SLAM system that separates objects by motion.
  \item StaticFusion (SF)~\cite{staticfusion_2018_ICRA}: A 3D reconstruction system that segments
        and ignores dynamic parts.
\end{itemize}

Note that Co-Fusion and \methodtitle{} are the only systems that
maintain multiple object models. The sequences in Table~\ref{tab:eval}
are roughly ordered by difficulty and latter rows exhibit an
increasing amount of dynamic motion. While \textit{f3s} abbreviates
\textit{freiburg3\_sitting}, \textit{f3w} stands for
\textit{freiburg3\_walking}.

Interestingly, ElasticFusion performs best in the presence of slight motion, even though it assumes
 static scenes. Our interpretation of this is that other methods label points as dynamic / outlier
that would still be beneficial for tracking, and hence show inferior performance.

Making use of context information proves to be especially useful in
highly dynamic scenes, or when the beginning of a scene is
difficult. These cases can be hard to tackle by energy minimisation,
whereas semantic segmentation results are shown to be robust.

\begin{figure} \includegraphics[width=\linewidth]{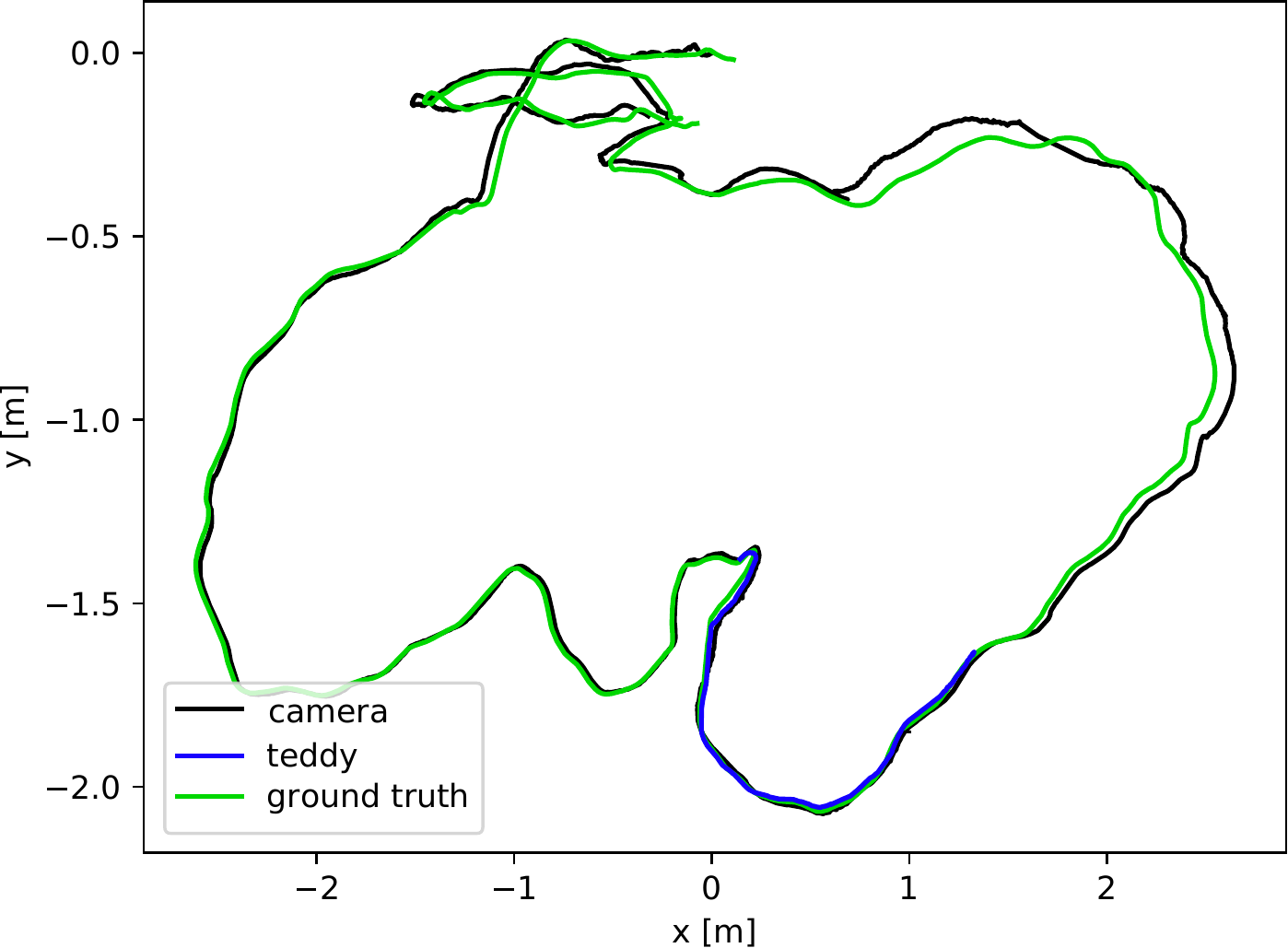} \caption{Comparison of camera and
object trajectories with ground-truth. The AT-RMSEs amount to $2.2$cm and $8.9$cm for the teddy bear and
camera trajectory, respectively. Because the bear occupies a significant proportion of the field of
view, tracking it independently affects the quality of the camera pose estimation. Treating the
object as part of the background would reduce the camera AT-RMSE to $7.2$cm.} \label{fig:f3long}
\end{figure}

Further, we reconstruct and track the teddy bear in sequence \textit{f3\_long\_office} independently
from the background motion. This way it is possible to compare the estimated object trajectory with
the ground-truth camera trajectory, as highlighted in Figure~\ref{fig:f3long}. The trajectory of the
bear is only available for a subsection of the sequence as it is out-of-view otherwise.

\subsubsection{Reconstruction}

\begin{figure}
  \centering
  \subfloat[Real object]{\includegraphics[width=.2\linewidth]{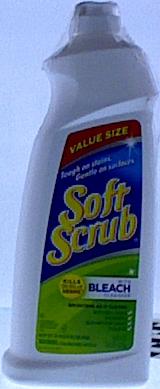}}~~
  \subfloat[3D model]{\includegraphics[width=.2\linewidth]{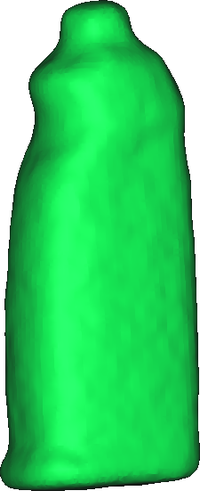}}~~
  \subfloat[Ours]{\includegraphics[width=.2\linewidth]{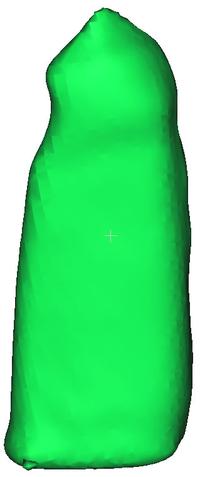}}~~
  \subfloat[Error]{\includegraphics[width=.2\linewidth]{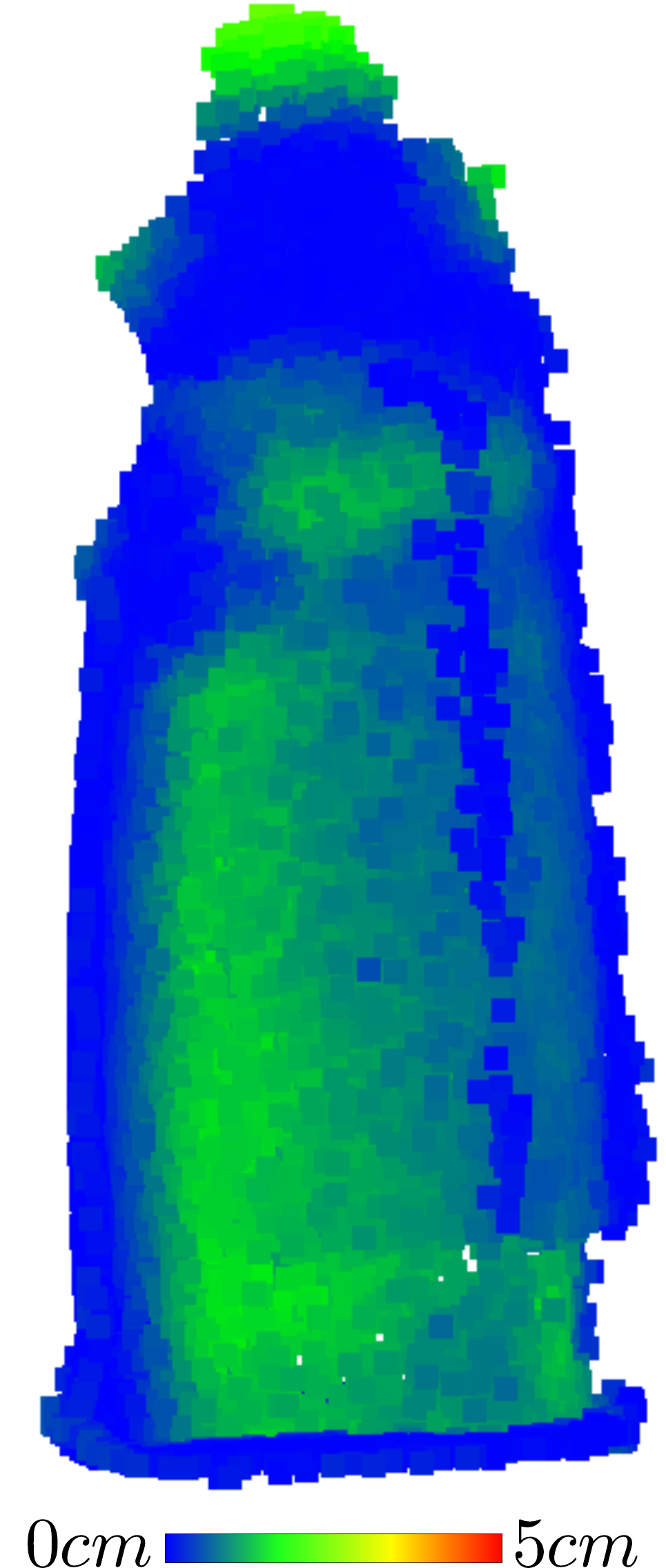}}
  \\
  \subfloat[Frame 500]{\includegraphics[width=.47\linewidth]{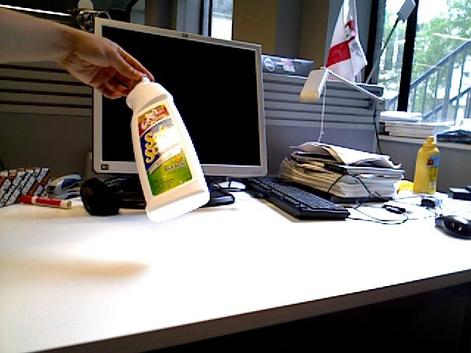}}~~
  \subfloat[Frame 950]{\includegraphics[width=.47\linewidth]{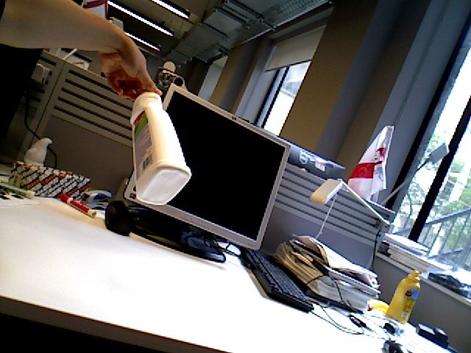}}
  \caption{Reconstruction of a bleach bottle from the YCB dataset. The average distance of a reconstructed surfel to a point on the ground-truth model is $7.0$mm with a standard deviation of $5.8$mm.}
\label{fig:reconstruction} \end{figure}

We conducted a quantitative evaluation of the quality of the 3D
reconstruction achieved by \methodtitle{} using objects from the YCB
Object and Model Set~\cite{YCBDataset}, a benchmark designed to
facilitate progress in robotic manipulation applications. The YCB set
provides physical daily life objects of different categories, which
are supplied to research teams, as well as a database with mesh models
and high-resolution RGB-D scans of the objects. We selected a ground
truth model from the dataset (a bleach bottle), and acquired a dynamic
sequence to quantitatively evaluate the errors in the 3D
reconstruction. Figure~\ref{fig:reconstruction} shows an image of the
object, the ground truth 3D model, our reconstruction and a heatmap
showing the 3D error per surfel. The average 3D error for the bleach
bottle was $7.0$mm with a standard deviation of $5.8$mm (where the GT
bottle is 250mm tall and 100mm across).

\subsubsection{Segmentation}

\begin{figure} \includegraphics[width=.98\linewidth]{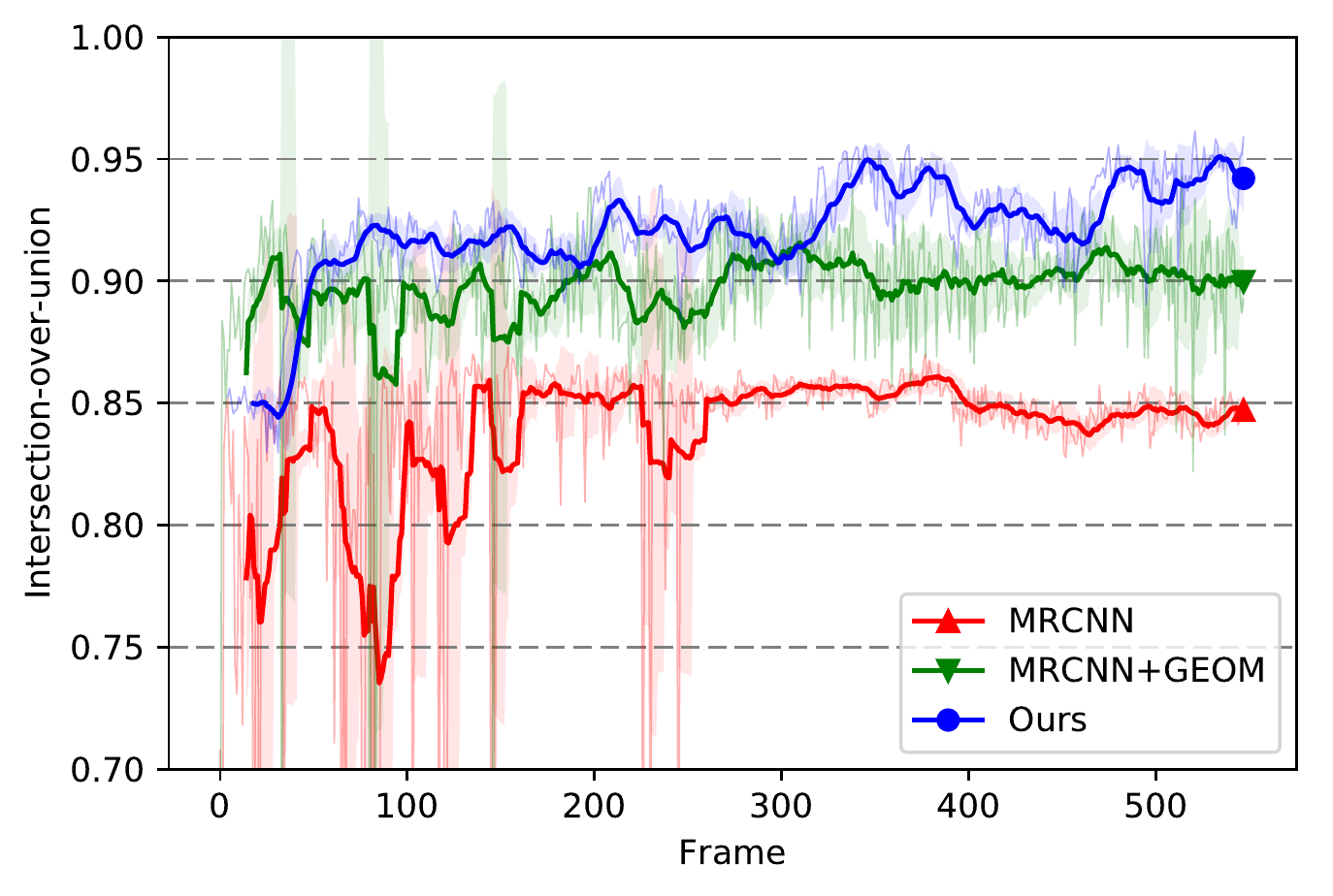}  \caption{Comparison of labelling
performance over time. Results of Mask-RCNN (MRCNN) and Mask-RCNN followed by our geometric
segmentation pipeline (MRCNN+GEOM) are frame-independent and variations in quality are only due to
changes in camera perspective. The blue graph (Ours) shows the intersection-over-union correlating
ground-truth 2D labels with the projection of the reconstructed 3D model.}
\label{fig:iou} \end{figure}

To assess the quality of the segmentation quantitatively we acquired a
$600$ frame long sequence and provided ground truth 2D annotations for
the masks of one of the objects (teddy). Figure~\ref{fig:iou} shows
the intersection over union (IoU) graphs for three different runs. The
IoU of the per-frame segmentation masks obtained with MaskRCNN only
and MaskRCNN combined with the geometric segmentation are shown in red
and blue respectively. The blue curve shows the IoU obtained using our
full method, where the object masks are obtained by reprojecting the
reconstructed 3D model. This graph shows how combining semantic and
geometric cues results in more accurate segmentations, but even better
results are achieved when maintaining temporally consistent 3D models
over the sequence through tracking and fusion.

\subsection{Qualitative results}

We tested \methodtitle{} on a variety of dynamic sequences, which show that it presents an effective
toolbox for different use cases.

\subsubsection{Grasping}

\begin{figure*}
  \center
   \def \imwidth {.32}
   \def \rowoffset {0.5cm}
    \textit{Skateboard} sequence \\ \vspace{-0.3cm}
    %\subfloat[Frame 100]{\includegraphics[width=\imwidth\textwidth]{frametable/X100}}
    \subfloat[Frame 300]{\includegraphics[width=\imwidth\textwidth]{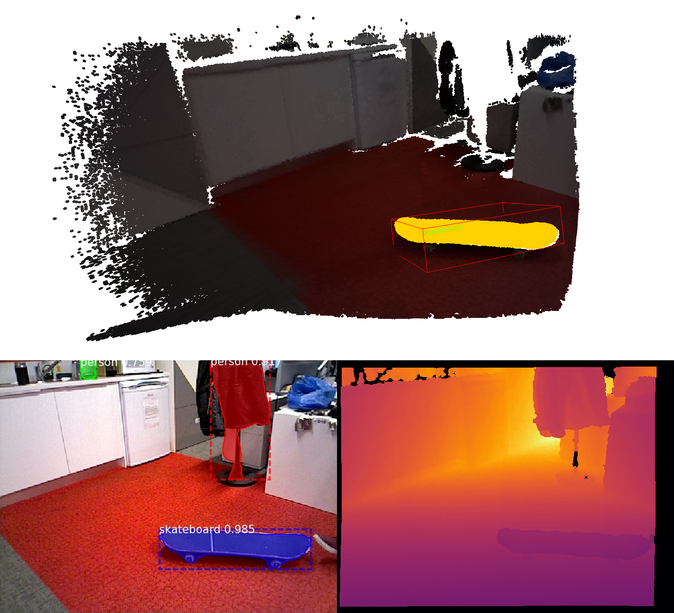}}
    \subfloat[Frame 325]{\includegraphics[width=\imwidth\textwidth]{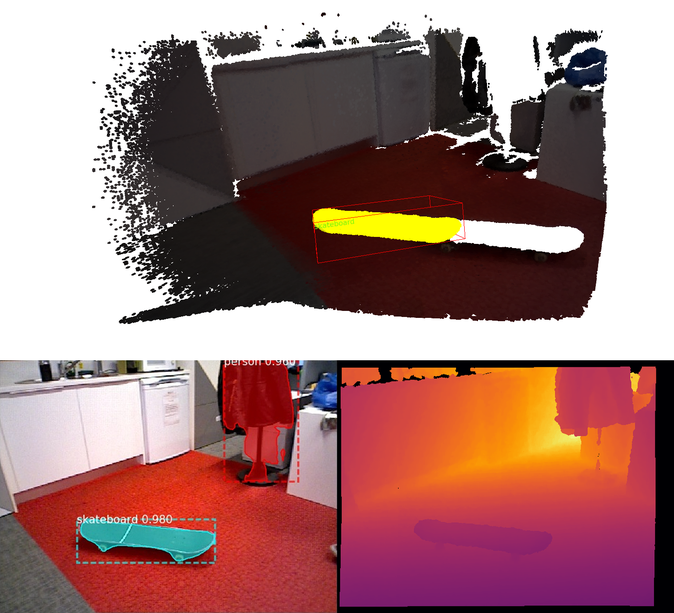}}
    \subfloat[Frame 350]{\includegraphics[width=\imwidth\textwidth]{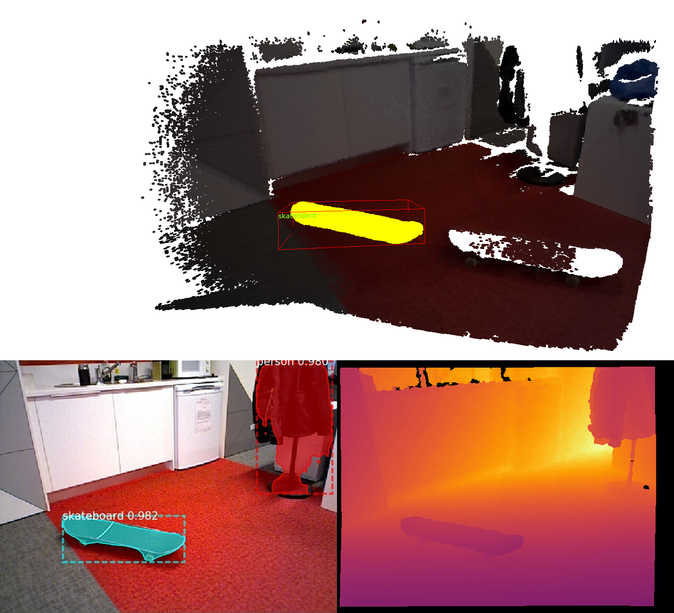}}
    \vspace{\rowoffset} \\
    \textit{Tidy-up} sequence \\ \vspace{-0.3cm}
    \subfloat[Frame 400]{\includegraphics[width=\imwidth\textwidth]{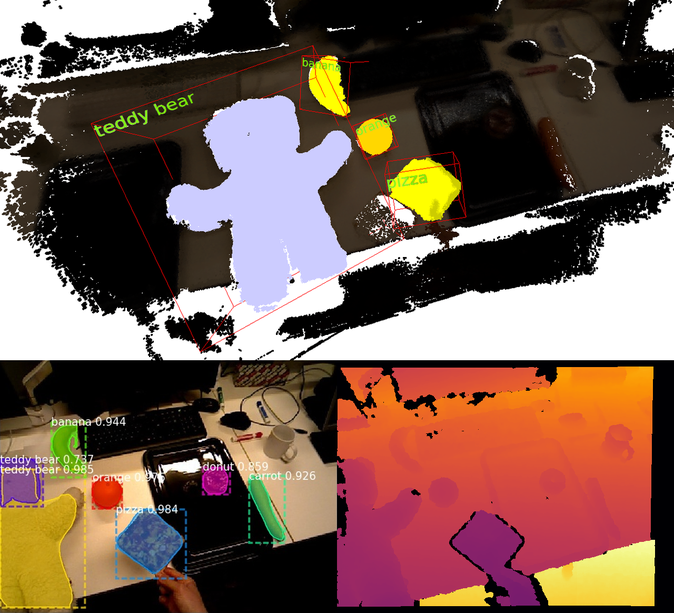}}
    %\subfloat[Frame 600]{\includegraphics[width=\imwidth\textwidth]{frametable/T600}}
    \subfloat[Frame 800]{\includegraphics[width=\imwidth\textwidth]{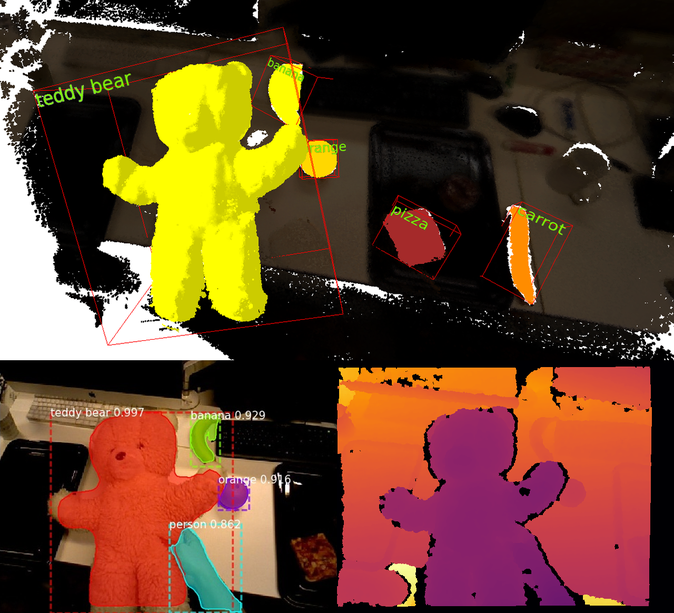}}
    \subfloat[Frame 1000]{\includegraphics[width=\imwidth\textwidth]{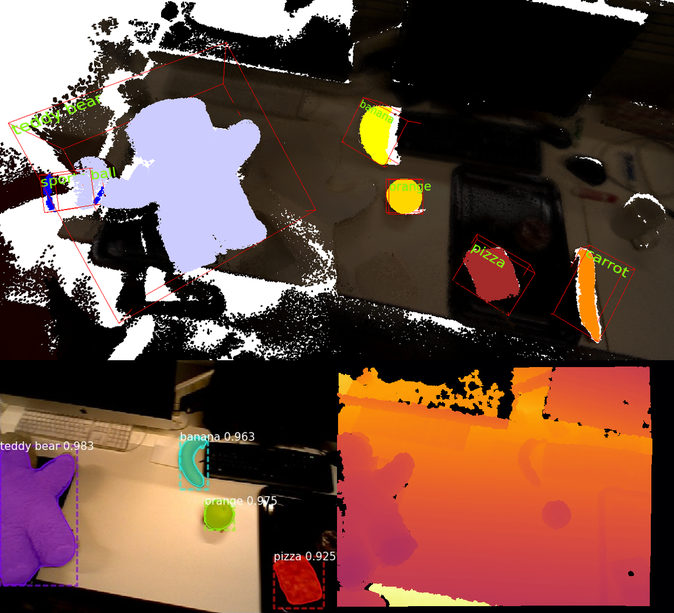}}
    \vspace{\rowoffset} \\
    \textit{Holding two objects} sequence \\ \vspace{-0.3cm}
    \subfloat[Frame 300]{\includegraphics[width=\imwidth\textwidth]{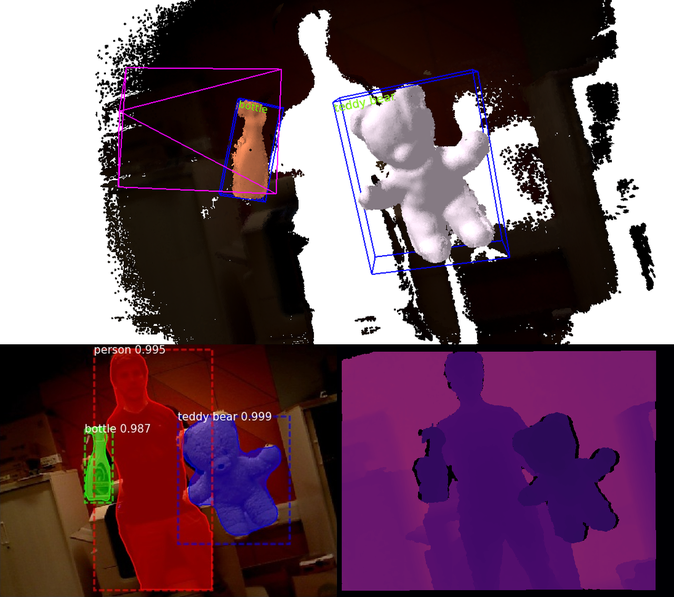}}
    \subfloat[Frame 500]{\includegraphics[width=\imwidth\textwidth]{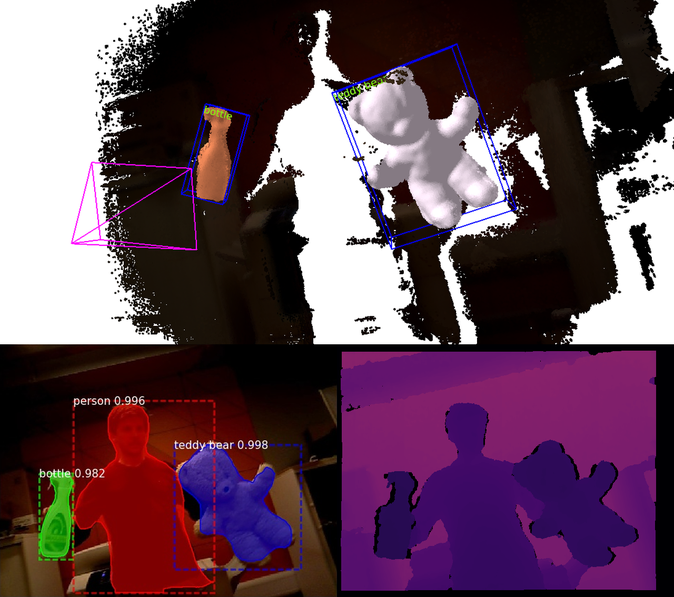}}
    %\subfloat[Frame 900]{\includegraphics[width=\imwidth\textwidth]{frametable/B900}}
    \subfloat[Frame 600]{\includegraphics[width=\imwidth\textwidth]{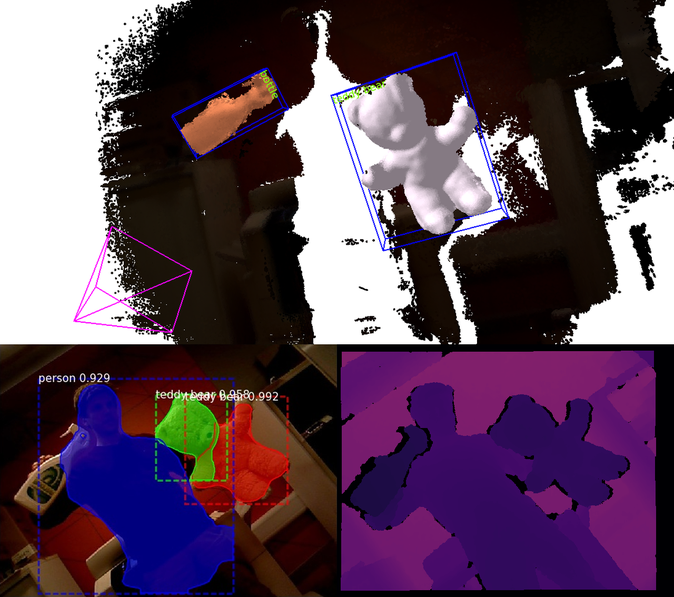}}

  \caption{Overview of evaluation sequences.}
  \label{fig:frametable}
\end{figure*}

A common but challenging task in robotics is to grasp objects. Aside from requiring sophisticated
actuators, a robot needs to identify grasping points on the correct object. \methodtitle{} is well
suited to provide the relevant data, as it detects and reconstructs objects densely. Further, and
in contrast to most other systems, it continues the tracking during interaction. If the appearance
of the actuator is known in advance or if a person interacts with objects, the neural network can be
trained to exclude these parts from the reconstruction. Figure~\ref{fig:frametable} shows a
timeline of frames that illustrate a grasping performance. In this example, the first 600 frames
were used to detect and model 5 objects in the scene, while tracking the camera. We implemented a
simple hand-detector that is used to recognise when an object is touched, and as soon as the person
interacts with the spray-bottle, the object is tracked reliably until it is placed back on the
table at frame 1100.

\begin{figure}
  \subfloat[Input frame 515]{\includegraphics[width=.44\linewidth]{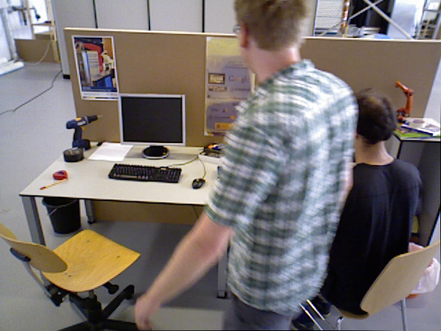}}~
  \subfloat[Reconstruction at frame 515]{\includegraphics[width=.54\linewidth]{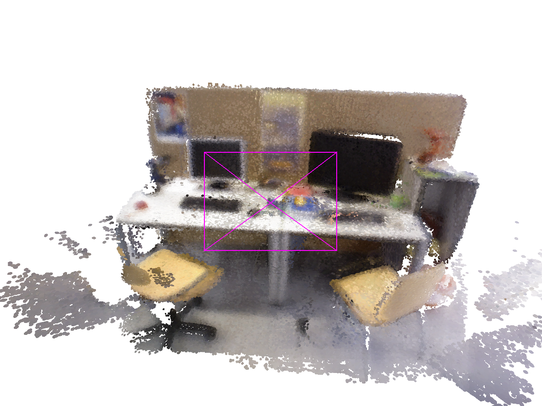}}
  \caption{Detecting persons allows \methodtitle{} to ignore them. In this challenging sequence (\textit{fr3\_walking\_halfsphere}), the reconstruction only contains static parts.}
  \label{fig:tum}
\end{figure}

\begin{figure}
  \centering
  \subfloat[Semantic reaction]{\includegraphics[width=0.32\linewidth]{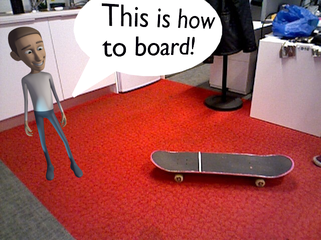}}~
  \subfloat[Object interaction]{\includegraphics[width=0.32\linewidth]{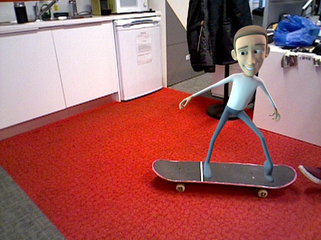}}~
  \subfloat[Obeying dynamics]{\includegraphics[width=0.32\linewidth]{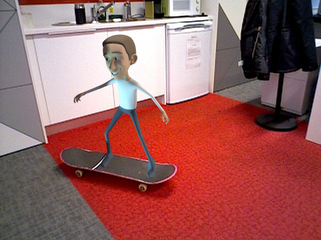}}
  \caption{AR application that shows a virtual character interacting with the scene.}
  \label{fig:skateboard}
\end{figure}

\begin{figure}
  \centering
  \subfloat[Showing estimated calories for a banana]{\includegraphics[width=0.49\linewidth]{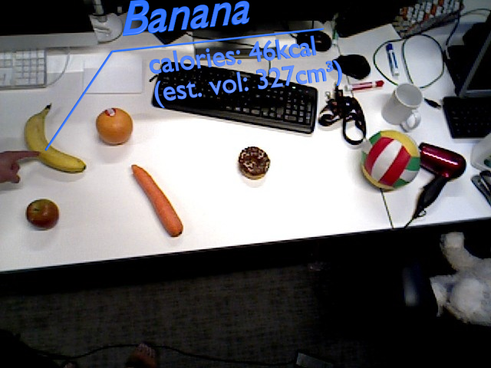}}~
  \subfloat[Showing estimated calories for a carrot]{\includegraphics[width=0.49\linewidth]{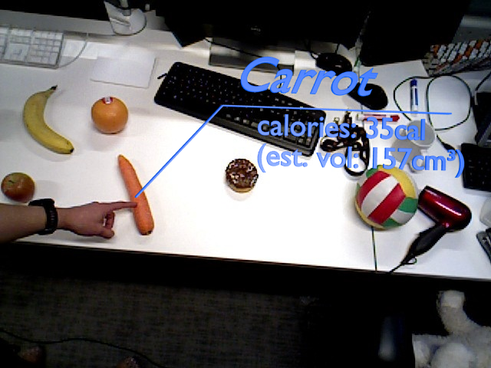}}
  \\
  \subfloat[3D reconstruction]{\includegraphics[width=0.49\linewidth]{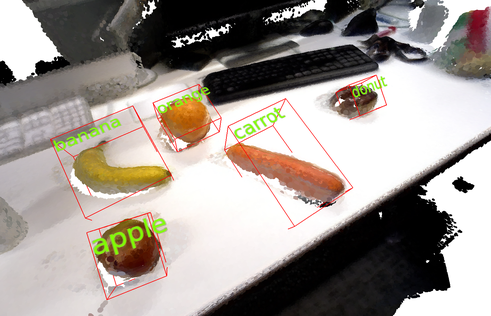}}~
  \subfloat[Object labels in 3D]{\includegraphics[width=0.49\linewidth]{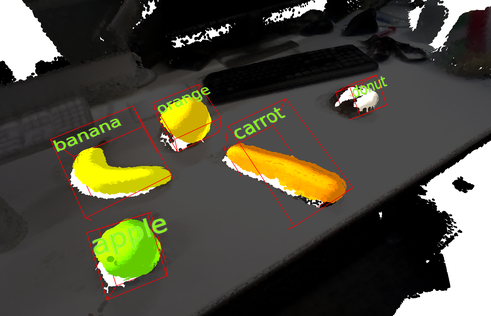}}
  \caption{AR application that estimates the calories of groceries.}
  \label{fig:calories}
\end{figure}

\subsubsection{Augmented reality}

Visual SLAM is a building block of many augmented reality systems and
we believe that adding semantic information enables new kinds of
applications. To illustrate that \methodtitle{} can be used for
augmented reality applications, we implemented demos that rely and
geometric as well as semantic data in dynamic scenes:

\textbf{Calories demo} This prototype aims at estimating the calories of an object-based on its
class and shape. By estimating body volumes, using simple primitive fitting, and providing a
database with \textit{calories per volume unit ratio}s for different classes, it is straightforward
to augment footage with the desired information. Experiments based on this prototype are shown in
Figure~\ref{fig:calories}.

\textbf{Skateboard demo} Another demo program presents a virtual character that actively reacts to
its environment. As soon as the skateboard appears in the scene the character jumps and remains on
it, as depicted in Figure~\ref{fig:skateboard}. Note that the character stays attached to the board
even after a person kicks it and sets it into motion. This requires accurate tracking of the
skateboard and camera at the same time.

\subsection{Performance}

The convolutional masking component runs asynchronously to the rest of
\methodtitle{} and requires a dedicated GPU. It operates at 5Hz, and
since it is blocking the GPU for long periods of time, we use another
GPU for the SLAM pipeline, which operates at $>$30Hz if a single model
is tracked. In the presence of multiple non-static objects, the
performance declines and results in a frame-rate of 20Hz for 3
models. Our test system is equipped with two Nvidia GTX Titan X and an
Intel Core i7, 3.5GHz.

%%%%%%%%%%%%%%%%%%%%%%%%%%%%%%%%%%%%%%%%%%%%%%%%%%%%%%%%%%%%%%%%%%%%%%%%%%%%%%%%%%%%%%%%%%%%%%%%%%%%
\section{CONCLUSIONS}

This paper introduced \methodtitle{}, a real-time visual SLAM system
that utilises semantic scene understanding to map and track multiple
objects. While inferring semantic labels from 2D image data, the
system maintains independent 3D models for each object instance and
for the background. We showed that \methodtitle{} can be used to
implement novel augmented reality applications or perform common
robotics tasks.

While \methodtitle{} makes meaningful progress towards achieving an
accurate, robust and general dynamic and semantic SLAM system, it
comes with limitations in the three main problems it addresses:
recognition, reconstruction and tracking.  Regarding the recognition,
\methodtitle{} can only recognise objects from classes on which
MaskRCNN~\cite{maskrcnn} has been trained (currently the 80 classes of
the MS-COCO dataset) and does not account for miss-classification of
object labels. Secondly, although \methodtitle{} can cope with the
presence of some non-rigid objects, such as humans, by removing them
from the map, tracking and reconstruction is limited to rigid
objects. Thirdly, tracking small objects with little geometric
information when no 3D model is available can result in
errors. Solving these limitations opens up opportunities for future
work.

%%%%%%%%%%%%%%%%%%%%%%%%%%%%%%%%%%%%%%%%%%%%%%%%%%%%%%%%%%%%%%%%%%%%%%%%%%%%%%%%%%%%%%%%%%%%%%%%%%%%

\begin{figure*}
\centering
\hspace{-1cm}
\begin{tabular}{l l l|l}
  \subfloat[Frame 300]{\includegraphics[width=0.24 \linewidth]{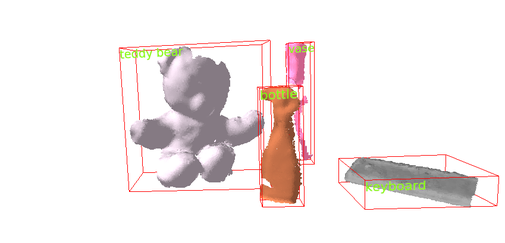}\hspace{-0.4cm}} &
  \subfloat[Frame 600]{\includegraphics[width=0.24 \linewidth]{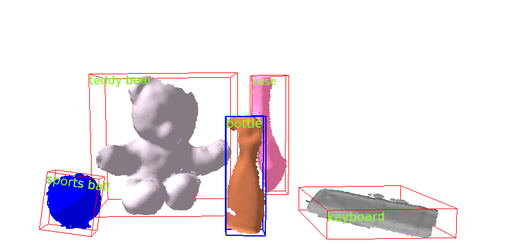}\hspace{-0.4cm}} &
  \subfloat[Frame 800]{\includegraphics[width=0.24 \linewidth]{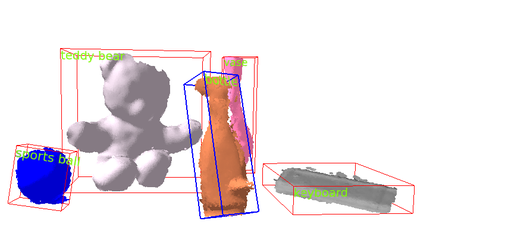}\hspace{-0.8cm}} &
  \subfloat[Reconstruction]{\includegraphics[width=0.2 \linewidth]{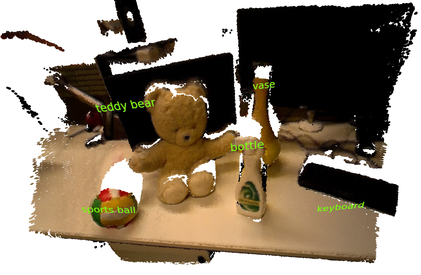}}
  \\
  \subfloat[Frame 900]{\includegraphics[width=0.24 \linewidth]{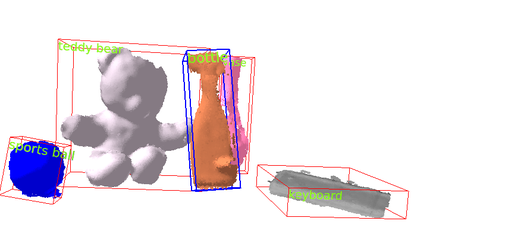}\hspace{-0.4cm}} &
  \subfloat[Frame 1000]{\includegraphics[width=0.24 \linewidth]{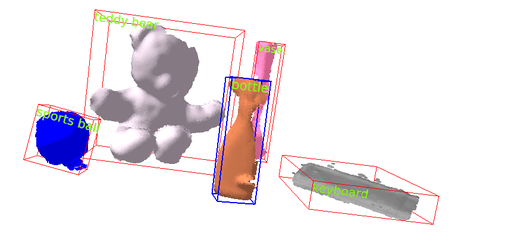}\hspace{-0.4cm}} &
  \subfloat[Frame 1160]{\includegraphics[width=0.24 \linewidth]{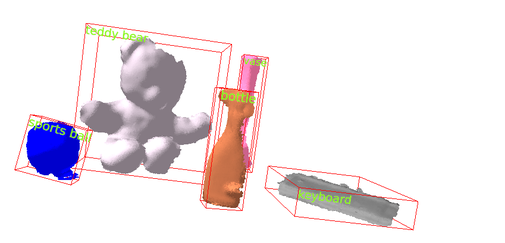}\hspace{-0.8cm}} &
  \subfloat[Normals]{\includegraphics[width=0.2 \linewidth]{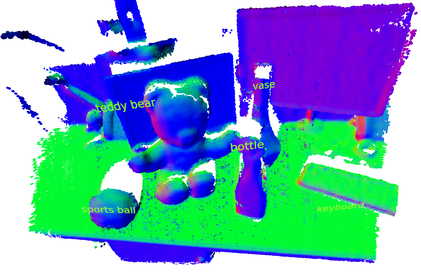}}
\end{tabular}
 \caption{A series of 6 frames, illustrating the recognition, tracking and mapping capabilities of \methodtitle{}. While a keyboard (grey), vase (pink), teddy-bear (white) and spray-bottle (orange) were detected from the beginning, the ball (blue) appeared between frame 300 and 600. The right hand side shows the reconstruction and estimated normals. The spray-bottle was moved by a person between frame 600 and 1000, but \methodtitle{} explicitly avoided to reconstruct person-related geometry.
 }
\end{figure*}


%%%%%%%%%%%%%%%%%%%%%%%%%%%%%%%%%%%%%%%%%%%%%%%%%%%%%%%%%%%%%%%%%%%%%%%%%%%%%%%%%%%%%%%%%%%%%%%%%%%%

% {\small
% \bibliographystyle{ieee}
% \bibliography{references}
% }
%%%%%%%%%%%%%%%%%%%%%%%%%%%%%%%%%%%%%%%%%%%%%%%%%%%%%%%%%%%%%%%%%%%%%%%%%%%%%%%%%%%%%%%%%%%%%%%%%%%%
\acknowledgments{
This work has been supported by the SecondHands project, funded from
the EU Horizon 2020 Research and Innovation programme under grant
agreement No 643950.}
\vspace{3cm}
\bibliographystyle{plain}
\bibliography{references}

\end{document}